\documentclass[pmlr]{jmlr} 
\newcommand{\tagdsmode}{proceedings}
\makeatletter
\newcommand{\tagdssubmission}{submission}
\newcommand{\tagdsproceedings}{proceedings}

\ifx\tagdsmode\tagdsproceedings
\else\ifx\tagdsmode\tagdssubmission
  \def\ps@jmlrtps{%
    \let\@mkboth\@gobbletwo
    \def\@oddhead{\scriptsize Under Review at the 2nd Conference on Topology, Algebra, and Geometry in Data Science\hfill}%
    \let\@evenhead\@oddhead
    \def\@oddfoot{}%
    \let\@evenfoot\@oddfoot
  }
\else
  \def\ps@jmlrtps{%
    \let\@mkboth\@gobbletwo
    \def\@oddhead{}%
    \let\@evenhead\@oddhead
    \def\@oddfoot{}%
    \let\@evenfoot\@oddfoot
  }
\fi\fi
\makeatother

\usepackage{longtable}
\usepackage{booktabs}
\usepackage[load-configurations=version-1]{siunitx} 

\usepackage[T1]{fontenc}
\usepackage[expansion=false]{microtype} 
\usepackage{subcaption}     
\usepackage{mathtools}
\usepackage{multirow}       
\usepackage{xspace}         
\usepackage{listings}
\usepackage{enumitem}
\setlist{nosep} 
\usepackage{float}

\hypersetup{colorlinks=true,urlcolor=blue,citecolor=blue,linkcolor=blue}

\usepackage[capitalize,noabbrev]{cleveref}

\crefname{lstlisting}{code}{codes}
\Crefname{lstlisting}{Code}{Codes}

\crefname{algocf}{Algorithm}{Algorithms}
\Crefname{algocf}{Algorithm}{Algorithms}

\lstdefinestyle{pyalgo}{
  language=Python,
  basicstyle=\ttfamily\small,
  keywordstyle=\color{blue},
  commentstyle=\color{gray},
  stringstyle=\color{teal},
  frame=single,
  columns=fullflexible,
  keepspaces=true,
  showstringspaces=false,
}

\theorembodyfont{\upshape}
\theoremheaderfont{\scshape}
\theorempostheader{:}
\theoremsep{\newline}

\usepackage{amsmath,amsfonts,bm}

\newcommand*{\rom}[1]{%
\textup{\uppercase\expandafter{\romannumeral#1}}%
}

\def\eqref#1{equation~\ref{#1}}

\def\1{\bm{1}}

\def\muon{\texttt{Muon}\xspace}

\def\adam{\texttt{Adam}\xspace}
\def\adamw{\texttt{AdamW}\xspace}
\def\scaledgd{\texttt{ScaledGD}\xspace}

\def\shampoo{\texttt{Shampoo}\xspace}

\def\mB{{\bm{B}}}

\def\mG{{\bm{G}}}

\def\mM{{\bm{M}}}

\def\mO{{\bm{O}}}
\def\mP{{\bm{P}}}
\def\mQ{{\bm{Q}}}
\def\mR{{\bm{R}}}

\def\mU{{\bm{U}}}
\def\mV{{\bm{V}}}
\def\mW{{\bm{W}}}
\def\mX{{\bm{X}}}

\def\mLambda{{\bm{\Lambda}}}

\def\mSigma{{\bm{\Sigma}}}

\DeclareMathAlphabet{\mathsfit}{\encodingdefault}{\sfdefault}{m}{sl}
\SetMathAlphabet{\mathsfit}{bold}{\encodingdefault}{\sfdefault}{bx}{n}

\DeclareMathOperator*{\argmax}{arg\,max}
\DeclareMathOperator*{\argmin}{arg\,min}

\DeclareMathOperator{\Tr}{Tr}

\newcommand{\cG}{\mathcal{G}}

\newcommand{\cP}{\mathcal{P}}

\newcommand{\bE}{\mathbb{E}}

\newcommand{\inner}[2]{\left\langle #1, #2 \right\rangle}

\newcommand{\rank}{\mathrm{rank}}

\newcommand{\diag}{\mathsf{diag}}

\newcommand{\R}{\mathbb{R}}
\newcommand{\loss}{L}
\newcommand{\stepsize}{\alpha}
\newcommand{\momentum}{\mu}
\newcommand{\cond}{\kappa}
\newcommand{\polar}{\operatorname{polar}}
\newcommand{\newtonschulz}{\operatorname{Newton\text{-}Schulz}}
\newcommand{\frob}[1]{\left\lVert #1 \right\rVert_{\mathrm{F}}}
\newcommand{\opnorm}[1]{\left\lVert #1 \right\rVert_{\mathrm{op}}}

\jmlrvolume{334}
\jmlryear{2026}
\jmlrworkshop{Topology, Algebra, and Geometry in Data Science}

\title[Muon for Matrix Factorization]{Reassessing Muon for Matrix Factorization}

\renewcommand{\thefootnote}{\fnsymbol{footnote}}

\ifx\tagdsmode\tagdssubmission
\else
  \author{\Name{Ali Parviz}\footnotemark[2] \Email{alparviz@ucsd.edu} \\
   \Name{Gal Mishne}\footnotemark[2] \Email{gmishne@ucsd.edu}\\
   \Name{Alex Cloninger}\footnotemark[1]\footnotemark[2] \Email{acloninger@ucsd.edu}\\
         \AND
     \addr }
\fi

\begin{document}
\maketitle
\ifx\tagdsmode\tagdssubmission\else
  \footnotetext[1]{Department of Mathematics, UC San Diego.}
  \footnotetext[2]{Halicio\u{g}lu Data Science Institute, UC San Diego.}
\fi
\renewcommand{\thefootnote}{\arabic{footnote}}

\begin{abstract}
\muon has recently emerged as a
strong optimizer for large-scale deep learning, where it reshapes gradient
updates through approximate orthogonalization and has been reported to
outperform \adam and \adamw on large language model training. Its empirical
success has motivated a growing theoretical literature that interprets \muon as
steepest descent under the spectral norm. Yet it remains unclear which of
\muon's advantages stem from its update rule itself and which are artifacts of
the scale, architecture, and data of modern deep networks. In this work we
isolate the optimizer from these confounders by studying \muon on a simple,
well-understood, and spectrally structured problem: low-rank matrix
factorization. Through a controlled and systematically tuned comparison against
adaptive baselines, we find that \muon does \emph{not} consistently outperform
\adamw in this setting, and that several previously reported advantages are
sensitive to hyperparameter choices. Our results give a more nuanced picture of
when spectrum-aware orthogonalization helps, and argue for evaluating modern
optimizers on controlled problems in addition to end-to-end benchmarks. 
\end{abstract}

\section{Introduction}
\label{sec:intro}

Recent advances in large-scale optimization have introduced a class of
optimizers that explicitly exploit the matrix structure of gradient updates.
Among these, \muon (MomentUm Orthogonalized by Newton--Schulz) has emerged as a
promising alternative to standard first-order methods \citep{jordan2024muon}.
Rather than applying a momentum update directly, \muon transforms the update
direction through an approximate orthogonalization step based on Newton--Schulz
iterations, effectively reshaping the spectrum of the gradient. 

Empirically, \muon has demonstrated strong performance on modern deep learning
workloads, including GPT-style models, where it improves training efficiency and
in some cases outperforms widely used optimizers such as \adam and \adamw
\citep{jordan2024muon, liu2025muonscalablellmtraining,
shah2025practicalefficiencymuon}. Notably, large-scale studies report up to a
twofold speedup over \adamw in multi-billion-parameter language model training
\citep{shah2025practicalefficiencymuon}. These successes have motivated a growing
body of theoretical work that studies \muon through the lenses of spectral-norm
steepest descent and constrained optimization. Existing analyses establish
convergence guarantees under various simplifying assumptions, including exact
orthogonalization, simplified momentum dynamics, and locally quadratic models
\citep{pethick2025normconstrainedlmo, li2025noteconvergencemuon,
shen2025convergencemuon, chen2025muonspectralnorm, kovalev2025understandinggo,
riabinin2025gluon, gruntkowska2025ef21muon, sato2025muonconvergence,
nagashima2026improvedconvergenceratesmuon}, while more recent work studies
inexact Newton--Schulz iterations and how approximation errors propagate into
convergence guarantees \citep{shulgin2025beyondideal, anon2025newtonschulz,
lau2025polargrad}. 

Despite this progress, it remains unclear which aspects of \muon's advantage are intrinsic to the optimizer and which arise from the complexity of large-scale deep learning, where architectural, scale, and data-dependent effects are hard to disentangle. Existing analyses rarely show \emph{when} and \emph{why} \muon should
outperform classical optimizers in concrete, well-specified problems, tending to concern idealized settings or purely local properties
\citep{davis2025whenspectral, su2025isotropic}. We take a complementary
perspective and study \muon on low-rank matrix factorization: a simple yet fundamental problem that allows systematic, exhaustive evaluation and where \muon's spectral nature might be expected to help, since optimization is governed by well-characterized curvature and singular-value structure. Our goal is to characterize both the strengths and limitations of \muon and to identify regimes where it does or does not outperform standard optimizers such as \adam and \adamw.

This focus exposes a gap between expectation and behavior, which we organize
around three questions.

\begin{enumerate}[
  label=\textbf{Q\arabic*:},
  leftmargin=*,
  itemsep=0em
]
  \item \textbf{Does \muon offer any advantage in structured problems such as
  matrix factorization?} Matrix factorization has explicit low-rank structure and is associated with a well-behaved optimization landscape. If \muon genuinely exploits spectral
  properties of the gradient, this should translate into faster convergence. If
  it does not, what limits its effectiveness relative to adaptive methods such
  as \adamw?

    \item \textbf{Are \muon's benefits tied to specific problem structures?}  Which properties of an optimization problem make orthogonalized updates advantageous? Do their benefits stem from specific geometric or spectral features of the objective, or do they generalize across matrix factorization tasks with varying conditioning, over-parameterization, and constraints?

  \item \textbf{How does \muon compare to adaptive methods under
  ill-conditioning?} Matrix factorization is sensitive to conditioning,
  especially when singular values decay rapidly. Optimizers such as \adamw
  implicitly adapt to coordinate-wise scaling. Does \muon's spectral
  normalization compete with or conflict with such adaptivity, and under which
  regimes does one dominate?
\end{enumerate}

\noindent \textbf{Findings.}
Contrary to its strong performance in large-scale models, Muon's advantage in these canonical settings is problem-dependent. On low-rank factorization and matrix completion, its reported gains disappear under equal tuning, with AdamW and GD matching or exceeding its performance. On NMF, however, Muon retains a consistent advantage, likely because its orthogonalized updates discourage redundant factors and promote more diverse representations. This dependence on problem structure suggests that Muon's broader success may also arise from properties specific to deep learning and underscores the value of controlled optimization benchmarks.

\section{Matrix factorization: symmetric and nonnegative variants}
\label{sec:problem}

We begin by fixing notation used throughout the paper. For a matrix $\mM$, let
$\sigma_i(\mM)$ denote its $i$-th largest singular value and $\sigma_{\min}(\mM)$
its smallest. We write $\opnorm{\mM}$ and $\frob{\mM}$ for the spectral and
Frobenius norms, respectively. For $k \le d$, let $\mathcal{O}_{d \times k}$ be
the set of matrices in $\R^{d \times k}$ with orthonormal columns. Given scalars
$a_1, \dots, a_d$, we write $\diag\{a_1, \dots, a_d\}$ for the diagonal matrix
with these entries. We discuss \muon~\citep{jordan2024muon} in \cref{app:muon-background} and defer a detailed discussion of related work to Appendix~\ref{app:related-work}.

\paragraph{Symmetric matrix factorization}
\label{sec:problem-mf}

We consider the symmetric matrix factorization problem
\begin{equation}
  \min_{\mU \in \R^{d \times k}}\quad
  f(\mU) \;=\; \frac{1}{4}\,\frob{\mU \mU^\top - \mM^\star}^2 ,
  \label{eq:mf}
\end{equation}
where $\mM^\star \in \R^{d \times d}$ is a rank-$r$ positive semidefinite matrix
and $\mU \in \R^{d \times k}$ is a factor with $k \ge r$ columns ($k = r$ is the
exactly parameterized case and $k > r$ is over-parameterized). The goal is to
recover $\mM^\star$ through a low-rank factorization $\mU\mU^\top$ by solving
\eqref{eq:mf}.
We assume $\mM^\star$ admits the eigendecomposition
$\mM^\star = \mV^\star \mLambda^\star \mV^{\star\top}$, where
$\mLambda^\star = \diag\{\lambda_1^\star, \dots, \lambda_r^\star\}$ collects the
nonzero eigenvalues with $\lambda_1^\star \ge \cdots \ge \lambda_r^\star > 0$ and
$\mV^\star \in \R^{d \times r}$ is orthonormal. The condition number of
$\mM^\star$ is
\begin{equation}
  \cond \;\coloneqq\; \lambda_1^\star / \lambda_r^\star .
  \label{eq:condition-number}
\end{equation}
The symmetric problem~\eqref{eq:mf} is the positive semidefinite, factor-tied
instance of a broader family of low-rank factorization problems; we defer the
general bilinear factorization and the matrix completion variant to
Appendix~\ref{sec:problem-mf-variants}.

\paragraph{Nonnegative matrix factorization.}
A closely related variant constrains both the target and the factors to be
entrywise nonnegative. For a rank-$r$ target $\mM^\star \in \R^{d_1 \times d_2}$,
nonnegative matrix factorization (NMF) solves
\begin{equation}
  \min_{\substack{\mU \in \R_{\ge 0}^{d_1 \times k}\\[2pt]
                  \mV \in \R_{\ge 0}^{d_2 \times k}}}\quad
  \frac{1}{2}\,\frob{\mU \mV^\top - \mM^\star}^2 ,
  \label{eq:nmf}
\end{equation}
where $\R_{\ge 0}$ denotes the nonnegative reals and $k \ge r$. NMF is the
nonnegativity-constrained special case of the general bilinear factorization
\eqref{eq:bmf} (Appendix~\ref{sec:problem-mf-variants}). Moreover, We analyze the exact softplus and its Taylor surrogate as nonlinear matrix factorizations and show the truncation imposes a rank cap that explains the high-rank gap in Fig.~\ref{fig:kernel-lr-sweep} (Appendix.~\ref{app:softplus-mf}).

\begin{table}[t]
\centering
\small
\caption{Learning-rate-tuned final loss (geometric mean over $3$ seeds; lower is
better). The best optimizer per row is in \textbf{bold}. The tuned ranking varies
across problems and, within each problem, differs from the ranking at a fixed
default learning rate.}
\label{tab:tuned}
\begin{tabular}{llcccc}
\toprule
Problem & Setting & Muon & AdamW & GD & SignGD \\
\midrule
\multirow{5}{*}{Factorization}
 & $\kappa{=}1$   & $2.3\!\times\!10^{-8}$  & $1.9\!\times\!10^{-13}$ & $\mathbf{1.6\!\times\!10^{-13}}$ & $2.8\!\times\!10^{-7}$ \\
 & $\kappa{=}5$   & $3.0\!\times\!10^{-7}$  & $1.9\!\times\!10^{-12}$ & $\mathbf{1.7\!\times\!10^{-12}}$ & $4.4\!\times\!10^{-6}$ \\
 & $\kappa{=}25$  & $3.0\!\times\!10^{-6}$  & $4.2\!\times\!10^{-11}$ & $\mathbf{3.7\!\times\!10^{-11}}$ & $6.1\!\times\!10^{-3}$ \\
 & $\kappa{=}125$ & $4.1\!\times\!10^{-7}$  & $1.4\!\times\!10^{-9}$  & $\mathbf{1.0\!\times\!10^{-9}}$  & $4.4\!\times\!10^{-1}$ \\
 & $\kappa{=}625$ & $9.4\!\times\!10^{-5}$  & $\mathbf{2.6\!\times\!10^{-7}}$ & $1.3\!\times\!10^{-1}$ & $1.2\!\times\!10^{1}$ \\
\midrule
\multirow{5}{*}{Completion ($\kappa$)}
 & $\kappa{=}1$   & $\mathbf{4.6\!\times\!10^{-16}}$ & $4.8\!\times\!10^{-16}$ & $4.5\!\times\!10^{-11}$ & $2.6\!\times\!10^{-15}$ \\
 & $\kappa{=}5$   & $5.7\!\times\!10^{-15}$ & $\mathbf{5.5\!\times\!10^{-15}}$ & $1.6\!\times\!10^{-9}$ & $2.4\!\times\!10^{-13}$ \\
 & $\kappa{=}25$  & $\mathbf{1.4\!\times\!10^{-13}}$ & $1.4\!\times\!10^{-13}$ & $4.7\!\times\!10^{-7}$ & $3.5\!\times\!10^{-5}$ \\
 & $\kappa{=}125$ & $3.3\!\times\!10^{-7}$ & $\mathbf{1.7\!\times\!10^{-7}}$ & $1.8\!\times\!10^{-2}$ & $2.2\!\times\!10^{-2}$ \\
 & $\kappa{=}625$ & $\mathbf{9.8\!\times\!10^{-3}}$ & $2.0\!\times\!10^{-2}$ & $3.4\!\times\!10^{-2}$ & $3.6\!\times\!10^{-2}$ \\
\midrule
\multirow{3}{*}{Completion (rank)}
 & $r{=}4$   & $5.7\!\times\!10^{-15}$ & $\mathbf{5.5\!\times\!10^{-15}}$ & $1.6\!\times\!10^{-9}$ & $2.4\!\times\!10^{-13}$ \\
 & $r{=}5$   & $\mathbf{4.5\!\times\!10^{-15}}$ & $1.8\!\times\!10^{-12}$ & $2.1\!\times\!10^{-6}$ & $4.9\!\times\!10^{-8}$ \\
 & $r{=}100$ & $2.1\!\times\!10^{-15}$ & $\mathbf{1.5\!\times\!10^{-15}}$ & $1.8\!\times\!10^{-14}$ & $3.2\!\times\!10^{-15}$ \\
\midrule
\multirow{3}{*}{NMF (uniform)}
 & $r{=}10$  & $\mathbf{1.5\!\times\!10^{-9}}$ & $6.1\!\times\!10^{-3}$ & $4.8\!\times\!10^{-1}$ & $8.7\!\times\!10^{-1}$ \\
 & $r{=}50$  & $4.5\!\times\!10^{-9}$ & $\mathbf{4.1\!\times\!10^{-9}}$ & $4.1\!\times\!10^{0}$ & $1.3\!\times\!10^{2}$ \\
 & $r{=}100$ & $\mathbf{8.9\!\times\!10^{-6}}$ & $7.8\!\times\!10^{-3}$ & $1.2\!\times\!10^{2}$ & $1.2\!\times\!10^{3}$ \\
\midrule
\multirow{3}{*}{NMF (decay)}
 & $r{=}10$  & $\mathbf{5.1\!\times\!10^{-7}}$ & $2.3\!\times\!10^{-3}$ & $3.8\!\times\!10^{-2}$ & $1.2\!\times\!10^{-1}$ \\
 & $r{=}50$  & $\mathbf{3.8\!\times\!10^{-9}}$ & $8.4\!\times\!10^{-4}$ & $1.0\!\times\!10^{0}$ & $1.5\!\times\!10^{1}$ \\
 & $r{=}100$ & $\mathbf{1.2\!\times\!10^{-6}}$ & $2.1\!\times\!10^{-3}$ & $6.3\!\times\!10^{0}$ & $3.4\!\times\!10^{2}$ \\
\bottomrule
\end{tabular}
\end{table}

\begin{figure*}[t]
\centering
\includegraphics[width=\textwidth]{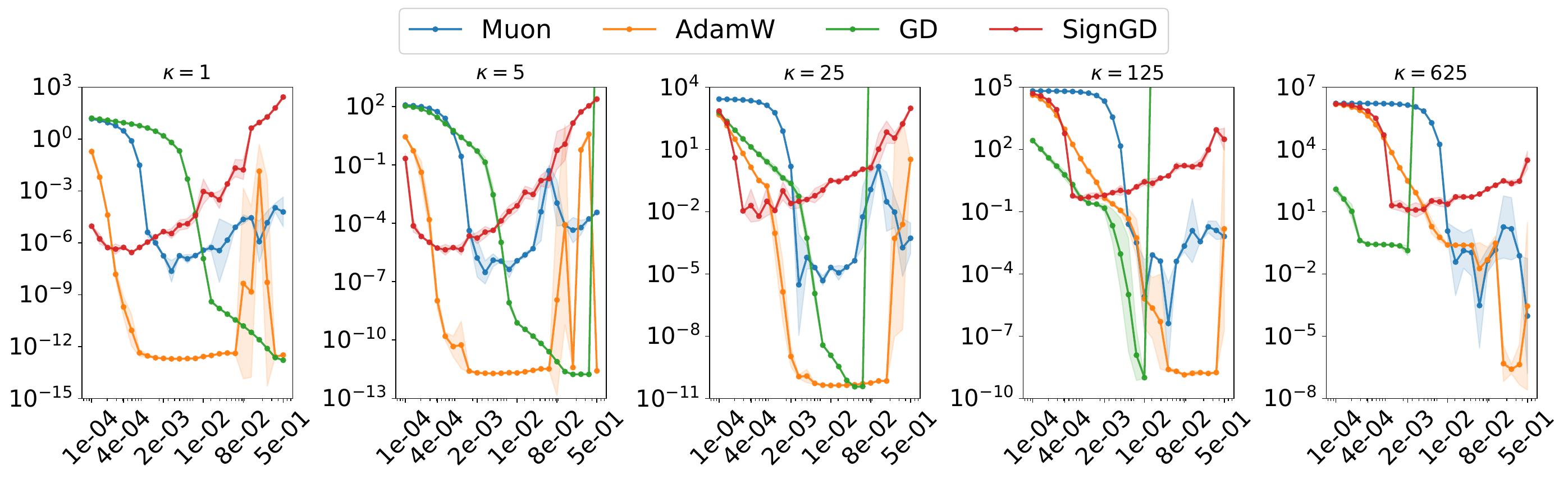}
\caption{\textbf{Low-rank factorization, conditioning sweep}
(target rank $r{=}15$, matched search rank $k{=}15$, dimension $d{=}100$).
Tuned final loss vs.\ learning rate for each condition number $\kappa$, from
well-conditioned ($\kappa{=}1$) to strongly ill-conditioned ($\kappa{=}625$)
(geometric mean over $3$ seeds, $\pm$log-std band). The target's $15$ singular
values are spaced linearly from $\kappa$ down to $1$, so $\kappa$ is exactly the
condition number. When tuned, AdamW and GD reach near machine precision; the
Muon configuration plateaus several orders higher.}
\label{fig:fact}
\end{figure*}

\begin{figure*}[t]
\centering
\includegraphics[width=\textwidth]{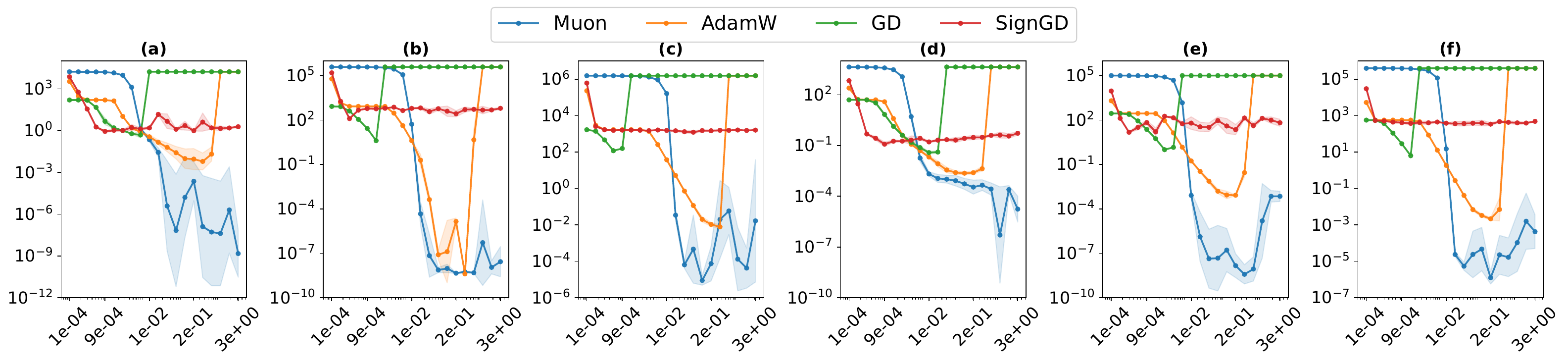}
\caption{\textbf{Non-negative factorization.} Final loss ($y$-axis, log scale;
geometric mean over $3$ seeds, shaded $\pm$one-log-standard-deviation band) versus
learning rate ($x$-axis, log scale). Panels \textbf{(a--c)} use the uniform
spectrum and \textbf{(d--f)} the decayed spectrum, at factor ranks
$r=10,50,100$ respectively. In both spectra, tuned Muon is the only method to fit
the target across ranks; AdamW trails by several orders and GD/SignGD fail.}
\label{fig:nmf-combined}
\end{figure*}

\begin{figure*}[t]
\centering
\includegraphics[width=\textwidth]{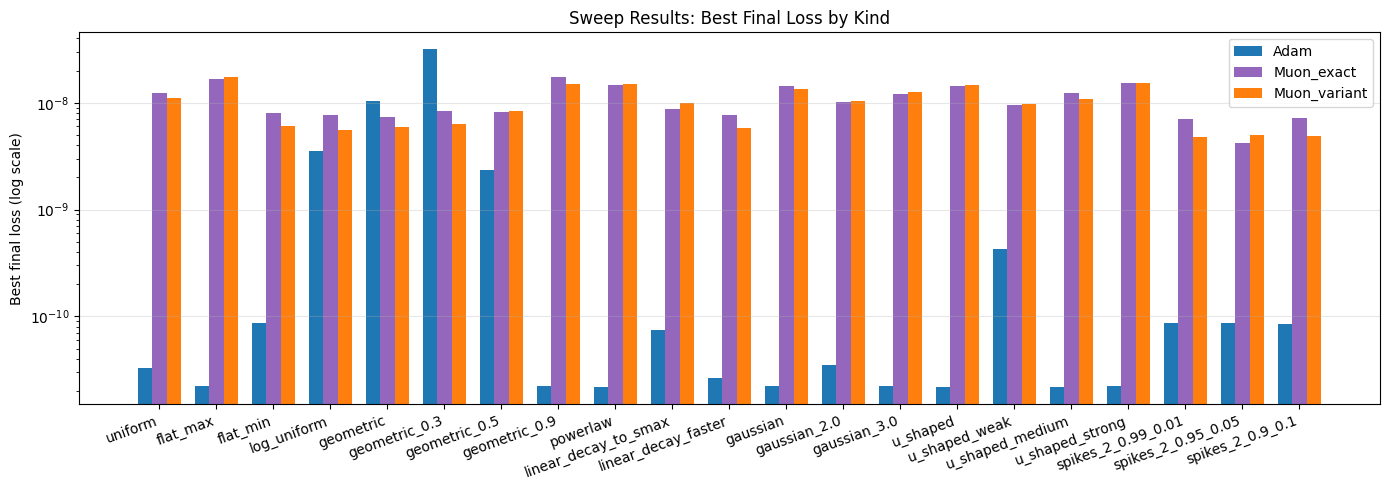}
\caption{\textbf{AdamW outperforms Muon variants across spectrum shapes}
(matrix factorization, target $m{\times}n = 100{\times}100$, rank $r{=}5$).
Final loss after LR tuning for each of the spectral
profiles at fixed $\kappa=10^{4}$; for every (profile, optimizer) pair we
report the best final loss over a logarithmic grid of learning rates, using a
common target instance and a shared initialization. \texttt{Muon\_exact}
(exact polar factor via SVD, no momentum) and \texttt{Muon\_variant}
(Newton--Schulz orthogonalization with Nesterov momentum, $\mu=0.95$) exhibit degraded relative performance on most spectral
shapes, whereas AdamW remains effective despite the extreme ill-conditioning.}
\label{fig:muon-spectrum}
\end{figure*}

\begin{figure*}[t]
  \centering
  \includegraphics[width=\textwidth]{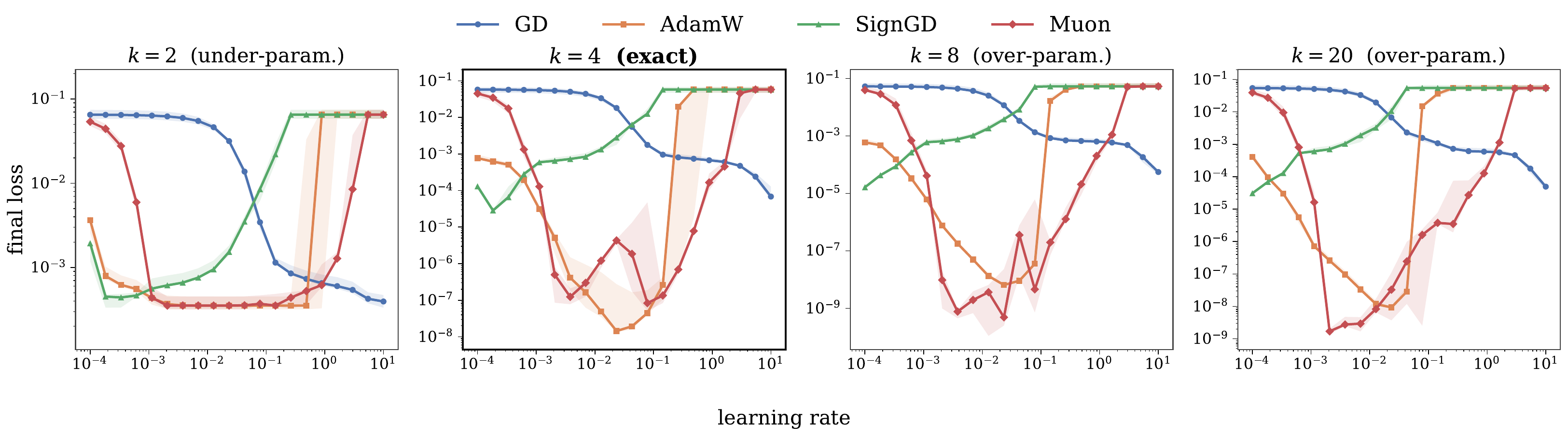}
  \caption{\textbf{Tensor-Train Noiseless regime} (true rank $r^\star=4$). Final loss versus
    learning rate for each optimizer, across search ranks $r\in\{2,4,8,20\}$
    spanning under-, exactly- ($r=r^\star$, bold panel), and over-parameterized
    settings. Solid lines are the median over $3$ seeds; shaded bands the
    min--max.}
  \label{fig:tt-clean}
\end{figure*}

\section{Experimental Design and Evaluation Protocol}
\label{sec:experiments}

We test the claim that optimizer comparisons in matrix factorization are highly
sensitive to hyperparameters, so that conclusions drawn from a single (often
default) configuration are not representative. Going beyond prior
work~\citep{ma2026preconditioningbenefitsspectralorthogonalization}, we evaluate
five matrix-recovery problems under a controlled protocol that holds model
capacity and parameterization fixed and varies only optimization-related
hyperparameters, spanning well-conditioned to severely \emph{ill-conditioned}
targets. Two further analyses isolate \emph{how} conditioning acts: the
sensitivity of each optimizer to the \emph{shape} of the eigenvalue distribution
at a fixed, extreme condition number (Section~\ref{sec:exp-spectrum}), and the
dynamics with which each recovers the target's spectral subspace
(Section~\ref{sec:exp-dynamics}).

\subsection{Problem settings}
\label{sec:exp-setup}

All problems use ambient dimension $d=100$ and a factorized parameterization
$\hat{M}=UV^{\top}$ with $U,V\in\mathbb{R}^{d\times r}$, optimized from a small
random initialization. We consider:

\begin{itemize}
\item \textbf{Low-rank factorization (conditioning).} A fully observed
      symmetric PSD target $M^{\star}=U^{\star}\mathrm{diag}(s)U^{\star\top}
      \in\mathbb{R}^{d\times d}$ of rank $r=15$, where $U^{\star}\in
      \mathbb{R}^{d\times r}$ has orthonormal columns (the thin $Q$-factor of a
      matrix with i.i.d.\ $\mathcal{N}(0,1)$ entries) and
      $s=(s_1,\dots,s_r)$ collects the nonzero eigenvalues of $M^{\star}$,
      spaced linearly between $s_1=\kappa$ and $s_r=1$:
      \[
        s_i \;=\; \kappa-\tfrac{i-1}{r-1}\,(\kappa-1),\qquad i=1,\dots,r .
      \]
      Since $U^{\star\top}U^{\star}=I_r$, the condition number of $M^{\star}$
      restricted to its range is exactly $s_1/s_r=\kappa$, and we sweep
      $\kappa\in\{1,5,25,125,625\}$ (at $\kappa=1$ the spectrum is flat and
      $M^{\star}$ is a rank-$r$ orthogonal projector). We solve for the
      symmetric factor $U\in\mathbb{R}^{d\times r}$ at matched search rank,
      minimizing $\tfrac{1}{4}\lVert UU^{\top}-M^{\star}\rVert_{F}^{2}$
      (Figure~\ref{fig:fact}).
\item \textbf{Matrix completion (conditioning).} A rank-$4$ symmetric PSD
        target with the same conditioning sweep, observed on a uniformly random
        $20\%$ of entries; loss
        $\tfrac{1}{2}\lVert\cP_\Omega(UV^{\top}-M^{\star})\rVert_{F}^{2}$, the
        squared error on the observed support (Figure~\ref{fig:compl-kappa}).
  \item \textbf{Matrix completion (over-parameterization).} The same completion
        task at fixed $\kappa=5$ and true rank $4$, sweeping the search rank
        $k\in\{4,5,100\}$ (matched, mildly, and heavily over-parameterized)
        (Figure~\ref{fig:compl-rank}).
\item \textbf{Non-negative factorization.} A target $M^{\star}$ with
      non-negative entries, factorized at rank $k\in\{10,50,100\}$ subject to
      an \emph{entrywise} non-negativity constraint on the factors,
      $U\in\mathbb{R}^{m\times k}_{\ge0}$ (i.e.\ $U_{ij}\ge0$ for all $i,j$;
      this is the componentwise order, not the positive-semidefinite one).
      The constraint is enforced by projected gradient descent: each optimizer
      step is followed by the Euclidean projection onto the non-negative
      orthant, $\Pi_{\ge0}(X)=\max(X,0)$, applied entrywise. We consider two
      target spectra: a \emph{uniform} spectrum and a \emph{decayed}
      (exponentially graded) spectrum, the latter removing the flat
      ``DC-component'' degeneracy of the uniform case, in which the leading
      non-negative component is nearly constant and all remaining directions
      carry equal weight (Figure~\ref{fig:nmf-combined}).
  \item \textbf{Tensor-train factorization (over-parameterization, noise).}
        A symmetric positive semidefinite target $M^{\star}=U^{\star}U^{\star\top}$
        recovered with untied cores $U,V$ in a noiseless ($r^{\star}=4$) and a
        noisy ($r^{\star}=30$) regime, sweeping the search rank across under-,
        exactly-, and over-parameterized values. Objective
        $\tfrac{1}{2}\lVert UV^{\top}-M^{\star}\rVert_{F}^{2}$; see
        Section~\ref{sec:problem-tt} and Figures~\ref{fig:tt-clean}
        and~\ref{fig:tt-noisy}.
\end{itemize}

We compare \textbf{Muon} (momentum $0.95$, Nesterov, $5$ Newton--Schulz
orthogonalization steps) against \textbf{AdamW}, plain
gradient descent (\textbf{GD}), and sign gradient descent (\textbf{SignGD}).

\subsection{Evaluation protocol}
\label{sec:exp-protocol}
We evaluate recovery with the \emph{normalized mean-squared error} (NMSE),
the squared reconstruction error relative to $\frob{\mM^\star}^2$, which makes
the metric scale-free and comparable across problems and conditioning regimes.
For every (problem, condition, optimizer) triple we sweep the learning
rate, the dominant axis and the usual source of misleading comparisons, over
a logarithmically spaced grid ($25$ points in $[10^{-4},5\times10^{-1}]$ for the
factorization and completion tasks; $20$ points in $[10^{-4},3.2]$ for NMF),
holding all other hyperparameters (momentum, and for \muon the Newton--Schulz
count $J$ and the orthogonalization coefficients) at standard values. Each
configuration runs for up to $3000$ iterations under an identical patience-based
learning-rate-decay schedule and is repeated over $3$ seeds. Because the
resulting NMSE values span roughly thirty orders of magnitude, we summarize each
configuration by the \emph{geometric mean} over seeds with a
$\pm$one-log-standard-deviation band, and compare optimizers by their
\emph{stable learning-rate range}, the band of step sizes over which each
converges to the NMSE floor.

\section{Experimental Results}
\label{sec:exp-results}
 
\noindent \textbf{A single learning rate is not a meaningful operating point.}
Within a fixed problem and optimizer, the final loss varies by a median of about
$11$ orders of magnitude across the learning-rate grid for the factorization and
completion tasks, and about $6$ orders for both NMF variants
(Figures~\ref{fig:fact}--\ref{fig:nmf-combined}). Performance is therefore dominated
by the learning rate, and any comparison made at one pre-selected value reflects
that choice as much as the optimizer itself.
 
\noindent \textbf{No shared learning rate is fair to all methods.}
The loss-minimizing learning rate differs \emph{across optimizers} by a median of
$1.5$--$3.1$ decades (largest for NMF), so evaluating all methods at a common
learning rate necessarily places at least one of them far from its own optimum.
This appears as a horizontal shift between each method's basin in every panel.
 
\noindent \textbf{Tuning reorders the methods, and the winner is problem-dependent.}
Table~\ref{tab:tuned} reports each optimizer's best (learning-rate-tuned) final
loss. In nearly all nineteen settings the ranking induced by a fixed
``default'' learning rate ($\approx 10^{-3}$) differs from the ranking obtained
once each method is tuned individually, and methods that look worst at the
default frequently become best after tuning. Under per-optimizer tuning no method
dominates universally: AdamW and GD win on plain low-rank factorization, Muon and
AdamW are comparable on matrix completion, and Muon wins clearly on both NMF
variants. The detailed per-setting comparisons, including the specific
default-vs-tuned reorderings, are deferred to
Appendix~\ref{sec:app-results-detail}.
 
\noindent \textbf{Ill-conditioning widens the gaps and shifts the winner.}
Conditioning is itself a hyperparameter of the problem, and it interacts strongly
with the optimizer. As $\kappa$ grows, every method's attainable loss degrades
and its optimal learning rate migrates upward (Figures~\ref{fig:fact},
\ref{fig:compl-kappa}). The effect is most dramatic for the non-adaptive methods
on factorization, where GD goes from the best-tuned method in the well-conditioned
regime to failing under extreme ill-conditioning while AdamW becomes best
(see Appendix~\ref{sec:app-results-detail} for the precise values). Thus
ill-conditioning does not merely raise the loss floor, it reorders the methods,
so that no fixed configuration summarizes behavior across the conditioning range.

\subsection{Sensitivity to spectrum shape under extreme ill-conditioning}
\label{sec:exp-spectrum}
 
Conditioning alone does not determine difficulty: at a \emph{fixed} condition
number, the convergence of first-order optimizers also depends on the
\emph{distribution} of eigenvalues within $[s_{\min}, s_{\max}]$. We therefore
hold the conditioning fixed at an extreme value and vary only the interior shape
of the spectrum, evaluating spectral profiles (defined in
Appendix~\ref{sec:app-spectral-profiles}). All families share the same
endpoints $(s_{\min}, s_{\max}) = (10^{-3}, 10)$ and hence the same, deliberately
extreme, condition number $\kappa = s_{\max}/s_{\min} = 10^{4}$; only the interior
shape differs. Despite this identical (and severe) ill-conditioning, the
distributions pose markedly different preconditioning challenges. As shown in
Figure~\ref{fig:muon-spectrum}, AdamW adapts effectively to heavy-tailed and
clustered spectra, whereas Muon underperforms across the majority of these
shapes, evidence that its spectral updates are less robust to specific
eigenvalue densities even when the overall condition number is held fixed.
Ill-conditioning is therefore not a single axis of difficulty: the same $\kappa$
can be easy or hard depending on how the eigenvalues are arranged, and the two
optimizers respond to that arrangement differently.

\begin{figure}[t]
\centering
\includegraphics[width=0.7\linewidth]{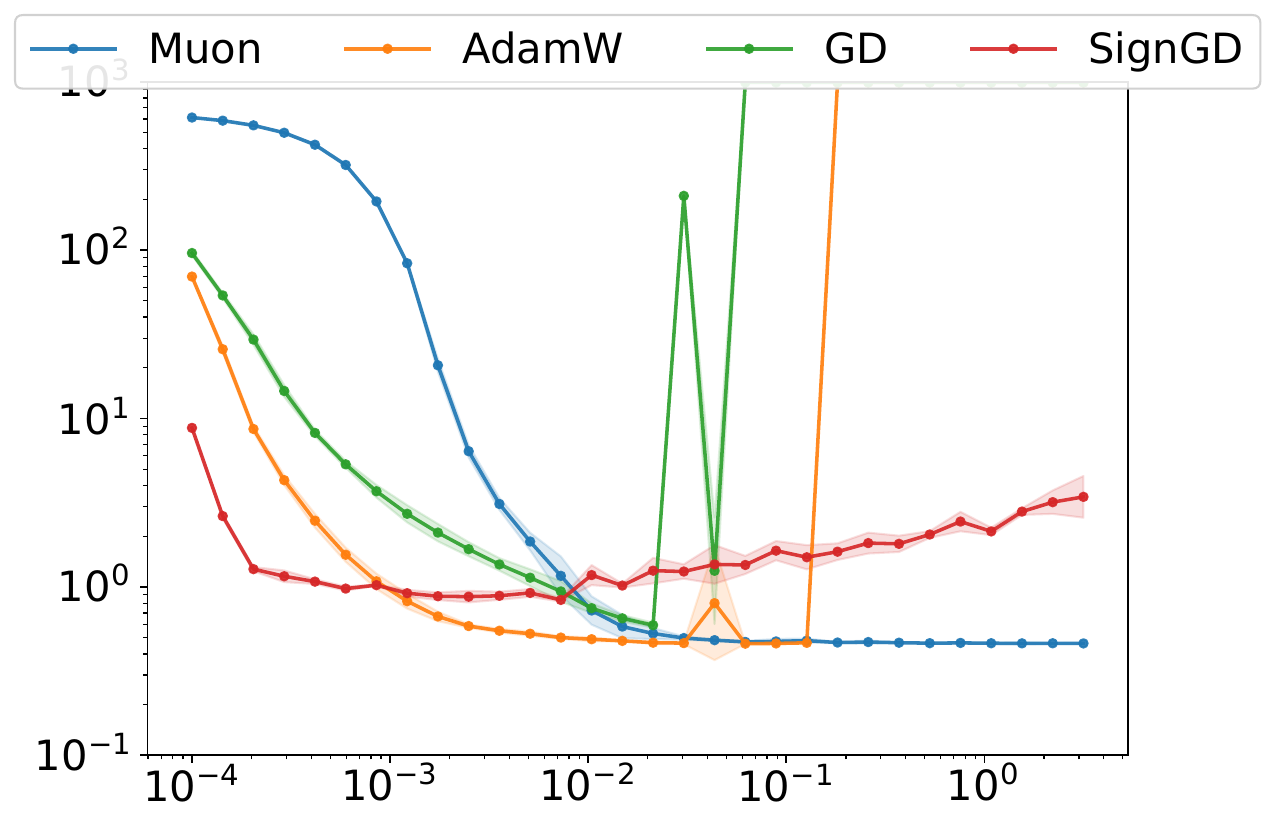}
\caption{\textbf{Gaussian-kernel NMF at a principled bottleneck.} Non-negative
factorization of a fixed Gaussian RBF kernel $K\in\mathbb{R}^{100\times100}$ at
search rank $r{=}24$ (the smallest rank capturing $95\%$ of $K$'s spectral
energy). On this smooth, well-behaved
landscape Muon's orthogonalized updates give no advantage: AdamW reaches the same
global floor as Muon.}
\label{fig:nmf-gaussian}
\end{figure}

\subsection{Gaussian-kernel factorization: a well-behaved landscape}
\label{app:gaussian-kernel}

To complement the synthetic NMF problems, we study the Gaussian radial-basis-function kernel. We sample
$N=100$ points in $\mathbb{R}^{5}$ and form $K_{ij}=\exp(-\lVert x_i-x_j\rVert^2/
2\sigma^2)$ with $\sigma=2$, giving a symmetric positive-semidefinite matrix with
strictly positive entries and a rapidly decaying spectrum. We factorize it as
$\hat{K}=UV^{\top}$ under a non-negativity constraint ($U,V\geq 0$, enforced by
projection after each step), minimizing $\tfrac{1}{2}\lVert UV^{\top}-K\rVert_F^2$.

Rather than fix the factor rank arbitrarily, we set it by a \emph{principled
bottleneck}: the smallest rank $r$ whose leading singular values capture at least
$95\%$ of $K$'s spectral energy, which yields $r=24$ for this target.  Each optimizer is tuned by a learning-rate sweep ($30$ points in
$[10^{-4},10^{0.5}]$), and every configuration is repeated over
three seeds that vary the initialization while holding the target fixed.

The result is shown in Figure~\ref{fig:nmf-gaussian}. On this well-behaved
landscape, Muon's advanced orthogonalization is unnecessary: standard AdamW
performs just as effectively, converging seamlessly to the same global floor as
Muon. This reinforces the study's central theme from the opposite direction, on
the adversarial NMF spectra Muon's structured updates help, but on a smooth, well-conditioned kernel target that advantage disappears and a well-tuned adaptive
baseline is equally good.

\subsection{Tensor-train factorization}
\label{sec:problem-tt}
We consider the tensor-train (TT) factorization of a matrix target. For an
order-$2$ tensor $\mM^\star \in \R^{d \times d}$, the TT decomposition contracts
two cores $\cG_1 \in \R^{d \times r}$ and $\cG_2 \in \R^{r \times d}$ with
boundary ranks $r_0 = r_2 = 1$ and a single internal rank $r_1 = r$,
\begin{equation}
  \mM^\star(i,j) \;=\; \sum_{\alpha=1}^{r}
    \cG_1(i,\alpha)\,\cG_2(\alpha,j) ,
  \qquad\Longleftrightarrow\qquad
  \mM^\star \;=\; \cG_1 \cG_2 ,
  \label{eq:tt-contraction}
\end{equation}
so the TT factorization coincides with the bilinear factorization of
Appendix~\ref{sec:problem-mf-variants} under $\mU = \cG_1$, $\mV^\top = \cG_2$,
with the lone TT-rank $r$ playing the role of the factorization rank. In our
setting the target $\mM^\star = \mU^\star \mU^{\star\top}$ is symmetric positive
semidefinite, but we do \emph{not} tie the cores: recovery uses untied factors
$\mU, \mV$, so the problem is the general bilinear factorization~\eqref{eq:bmf}
applied to a symmetric PSD target, retaining the $\mathrm{GL}_r$ imbalance
invariance of~\eqref{eq:bmf}. Recovery minimizes
\begin{equation}
  f(\mU, \mV) \;=\; \frac{1}{2}\,\frob{\mU\mV^\top - \mM^\star}^2 ,
  \qquad \mU \in \R^{d \times r}, \mV \in \R^{d \times r},
  \label{eq:tt-objective}
\end{equation}
the objective of~\eqref{eq:bmf} with search rank $r$ as the TT-rank.
 
We study \eqref{eq:tt-objective} in two regimes, a \emph{noiseless} regime
(true rank $r^\star = 4$, recovered exactly, loss limited only by convergence)
and a \emph{noisy} regime (true rank $r^\star = 30$ with additive observation
noise, loss floored at the statistical noise level), and within each we sweep
the search rank across under-, exactly-, and over-parameterized values; full
specifications are given in Appendix~\ref{sec:app-tt-detail}. The quantity of
interest is the \emph{stable learning-rate range} of each optimizer, the band of
step sizes over which~\eqref{eq:tt-objective} converges to its regime-dependent
floor, and how that band widens, narrows, or shifts as the parameterization
regime and the noise level change (Figures~\ref{fig:tt-clean}
and~\ref{fig:tt-noisy}). Beyond low-rank recovery, this factorized objective serves as a minimal model of the product parameterizations ubiquitous in deep learning, a connection we develop in Appendix~\ref{sec:tt-deep-learning}. We also study the effect of factorization depth in Appendix~\ref{sec:tt-depth-summary}.

\section{Conclusion}
\label{sec:conclusion}

In this work, we revisit optimizer comparisons for matrix factorization and show that conclusions are highly sensitive to hyperparameter selection and problem conditioning. Systematic learning-rate sweeps reveal that rankings reported under single configurations often do not hold: well-tuned AdamW and gradient descent frequently match or outperform Muon on standard low-rank factorization tasks, while Muon’s advantages are mostly limited to NMF and some highly ill-conditioned completion problems. We further find that optimizer performance depends on both conditioning severity and spectral structure, altering method rankings across problem instances. These results underscore the need to evaluate optimizers across tuned hyperparameter ranges and conditioning regimes rather than relying on default settings or isolated benchmarks.

\section*{Acknowledgements}
We thank Tianhao Wang for providing helpful references and for the discussions that assisted in developing the experimental setup. Ali Parviz was funded by NSF CIF-2403452.



\bibliography{references}

\newpage
\appendix
\onecolumn

\section{Background: The Muon Optimizer}
\label{app:muon-background}

We briefly recall the \muon update and the spectral viewpoint that motivates it.
Consider a weight matrix $\mW \in \R^{m \times n}$ with loss gradient
$\mG_t = \nabla_{\mW}\loss(\mW_t)$ at iteration $t$. \muon maintains a momentum
buffer $\mB_t$ and applies an \emph{orthogonalized} update:
\begin{align}
  \mB_t &= \momentum\, \mB_{t-1} + \mG_t, \label{eq:muon-momentum}\\
  \mO_t &= \newtonschulz(\mB_t) \;\approx\; \polar(\mB_t), \label{eq:muon-orth}\\
  \mW_{t+1} &= \mW_t - \stepsize\, \mO_t. \label{eq:muon-update}
\end{align}
Here $\stepsize > 0$ is the step size and $\momentum \in [0,1)$ the momentum
coefficient. The orthogonalization step \eqref{eq:muon-orth} replaces the
momentum direction by (an approximation of) its \emph{polar factor}.

\noindent \textbf{Polar factor and spectral steepest descent.}
Let $\mB_t = \mP \mSigma \mQ^\top$ be a singular value decomposition, with
$\mP \in \mathcal{O}_{m \times \rho}$, $\mQ \in \mathcal{O}_{n \times \rho}$, and
$\mSigma = \diag\{\sigma_1, \dots, \sigma_\rho\}$, where $\rho = \rank(\mB_t)$.
The polar factor is
\begin{equation}
  \polar(\mB_t) \;=\; \mP \mQ^\top,
  \label{eq:polar}
\end{equation}
which sets every nonzero singular value to one while preserving the singular
vectors. Equivalently, $\mP\mQ^\top$ is the solution of the spectral-norm linear
minimization oracle,
\begin{equation}
  \mP \mQ^\top \;=\; \argmax_{\opnorm{\mX} \le 1} \inner{\mB_t}{\mX},
  \label{eq:lmo}
\end{equation}
so the \muon direction \eqref{eq:muon-orth} is precisely the steepest-descent
direction with respect to the spectral norm. This is the sense in which \muon is
``spectrum aware'': it discards the magnitude of the momentum's singular values
and acts only along their directions.

\noindent \textbf{Newton--Schulz orthogonalization.}
Computing the SVD at every step is expensive, so \muon approximates the polar
factor with a fixed number of matrix-multiplication-only Newton--Schulz
iterations. Starting from the normalized matrix $\mX_0 = \mB_t / \frob{\mB_t}$,
one iterates a fixed odd polynomial
\begin{equation}
  \mX_{j+1} \;=\; a\,\mX_j \;+\; b\,\mX_j \mX_j^\top \mX_j
                  \;+\; c\,\big(\mX_j \mX_j^\top\big)^2 \mX_j,
  \qquad j = 0, 1, \dots, J-1,
  \label{eq:newton-schulz}
\end{equation}
with coefficients $(a,b,c)$ chosen so that the induced scalar map
$\sigma \mapsto a\sigma + b\sigma^3 + c\sigma^5$ pushes the singular values of
$\mX_j$ toward $1$. The quintic coefficients
$(a,b,c) = (3.4445, -4.7750, 2.0315)$ proposed by \citet{jordan2024muon} reach a
usable approximation in roughly $J = 5$ iterations; subsequent work designs
improved polynomials for this orthogonalization
\citep{amsel2025polarexpress, grishina2025chebyshevns, cesista2025muonoptcoeffs,
boumal2025polarpoly}. As $J \to \infty$ (with an exact map) the iteration
converges to $\polar(\mB_t)$, recovering the idealized update; in practice $J$ is
small and $\mO_t$ is only an approximate polar factor, which several recent
analyses account for explicitly \citep{shulgin2025beyondideal,
anon2025newtonschulz, lau2025polargrad}.

\section{Matrix Factorization Variants}
\label{sec:problem-mf-variants}
This appendix collects two generalizations of the symmetric matrix
factorization problem~\eqref{eq:mf}: the general bilinear (asymmetric)
factorization and the matrix completion setting in which only a subset of the
entries of the target is observed.

\paragraph{General bilinear factorization.}
The symmetric problem~\eqref{eq:mf} is a special case of the general
\emph{bilinear} (asymmetric) factorization, in which a possibly rectangular
target $\mM^\star \in \R^{d_1 \times d_2}$ of rank $r$ is recovered from the
product of two factors,
\begin{equation}
  \min_{\substack{\mU \in \R^{d_1 \times k}\\[2pt] \mV \in \R^{d_2 \times k}}}\quad
  f(\mU, \mV) \;=\; \frac{1}{2}\,\frob{\mU \mV^\top - \mM^\star}^2 ,
  \label{eq:bmf}
\end{equation}
with $k \ge r$. When $\mM^\star$ is symmetric positive semidefinite and the
factors are tied as $\mV = \mU$, \eqref{eq:bmf} reduces to~\eqref{eq:mf} up to a
constant rescaling of the objective. Unlike the symmetric case, the asymmetric
parameterization is invariant under the larger group of invertible
transformations $(\mU, \mV) \mapsto (\mU \mR, \mV \mR^{-\top})$ for any
$\mR \in \mathrm{GL}_k(\R)$, since $\mU \mV^\top = (\mU \mR)(\mV \mR^{-\top})^\top$.
This invariance permits an arbitrary imbalance between the factor norms and is
commonly controlled by a balancing regularizer
$\tfrac{1}{2}\bigl(\frob{\mU}^2 + \frob{\mV}^2\bigr)$ or by enforcing
$\mU^\top \mU = \mV^\top \mV$ along the trajectory.

\paragraph{Matrix completion.}
In many applications only a subset of the entries of $\mM^\star$ is observed.
Let $[n] \coloneqq \{1, \dots, n\}$ and let
$\Omega \subseteq [d_1] \times [d_2]$ denote the set of observed indices.
Define the sampling operator
$\cP_\Omega : \R^{d_1 \times d_2} \to \R^{d_1 \times d_2}$ by
\begin{equation}
  \bigl[\cP_\Omega(\mX)\bigr]_{ij}
  \;=\;
  \begin{cases}
    X_{ij}, & (i,j) \in \Omega,\\[2pt]
    0,      & \text{otherwise,}
  \end{cases}
  \label{eq:sampling-operator}
\end{equation}
which retains the observed entries and zeros out the rest. The factored matrix
completion problem is
\begin{equation}
  \min_{\substack{\mU \in \R^{d_1 \times k}\\[2pt] \mV \in \R^{d_2 \times k}}}\quad
  \frac{1}{2}\,\frob{\cP_\Omega\!\bigl(\mU \mV^\top - \mM^\star\bigr)}^2 .
  \label{eq:mc}
\end{equation}
Taking $\Omega = [d_1] \times [d_2]$ recovers the fully observed
problem~\eqref{eq:bmf}, while the symmetric variant tied to~\eqref{eq:mf}
additionally imposes $\mV = \mU$. Exact recovery in this setting requires the
ground-truth factors to be incoherent and the sample size $\lvert \Omega \rvert$
to be sufficiently large relative to the degrees of freedom $r(d_1 + d_2 - r)$.
Under the Bernoulli model in which each entry is observed independently with
probability $p$, the rescaled operator $p^{-1}\cP_\Omega$ is an unbiased estimator
of the identity, i.e.\ $\bE\bigl[p^{-1}\cP_\Omega(\mX)\bigr] = \mX$.

\begin{figure*}[t]
\centering
\includegraphics[width=\textwidth]{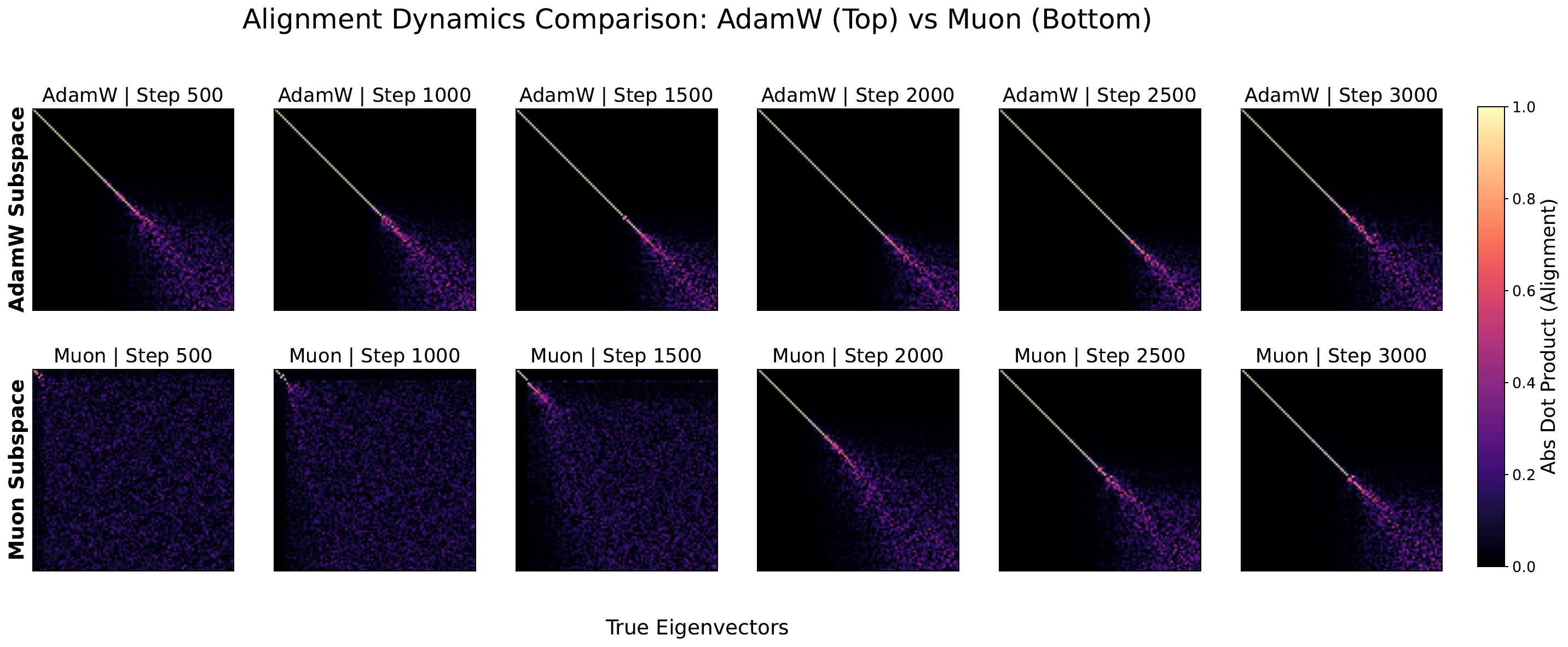}
\caption{\textbf{Spectral subspace recovery: Muon vs.\ AdamW.}
Absolute alignment $M_{ij}=\lvert\langle u_i, e_j\rangle\rvert$
(Eq.~\ref{eq:alignment}) between learned singular vectors $u_i$ and true
eigenvectors $e_j$ of $K$, every $500$ steps (AdamW top, Muon bottom). A bright,
correctly-ordered diagonal indicates faithful recovery; off-diagonal mass
reflects mixing. AdamW sharpens its diagonal steadily, while Muon stays
near-random for ${\sim}1500$ steps, then snaps into a more diffuse alignment.}
\label{fig:dynamics}
\end{figure*}

\section{Related Work}
\label{app:related-work}

\noindent \textbf{Spectrum-aware optimization and \muon.}
Much of the theoretical analysis of \muon builds on a perspective that
interprets modern optimizers such as \adam and \shampoo as instances of steepest
descent under non-Euclidean or norm-constrained geometries
\citep{bernstein2024old}. Within this view, \muon approximates a spectral-norm
steepest-descent direction via polar factorization, connecting it to a broader
class of methods that leverage matrix structure in gradients
\citep{pethick2025normconstrainedlmo, li2025noteconvergencemuon,
shen2025convergencemuon, chen2025muonspectralnorm}. Spectral transformations and
orthogonalization in optimization predate \muon, appearing in earlier work on
stochastic and preconditioned methods \citep{carlson2015stochasticB,
carlson2015stochastic, carlson2015preconditioned, tuddenham2022orthogonalising}.
A complementary line of work asks more directly when such spectral updates are
beneficial in deep learning \citep{davis2025whenspectral, su2025isotropic}.

\begin{figure*}[t]
\centering
\includegraphics[width=\textwidth]{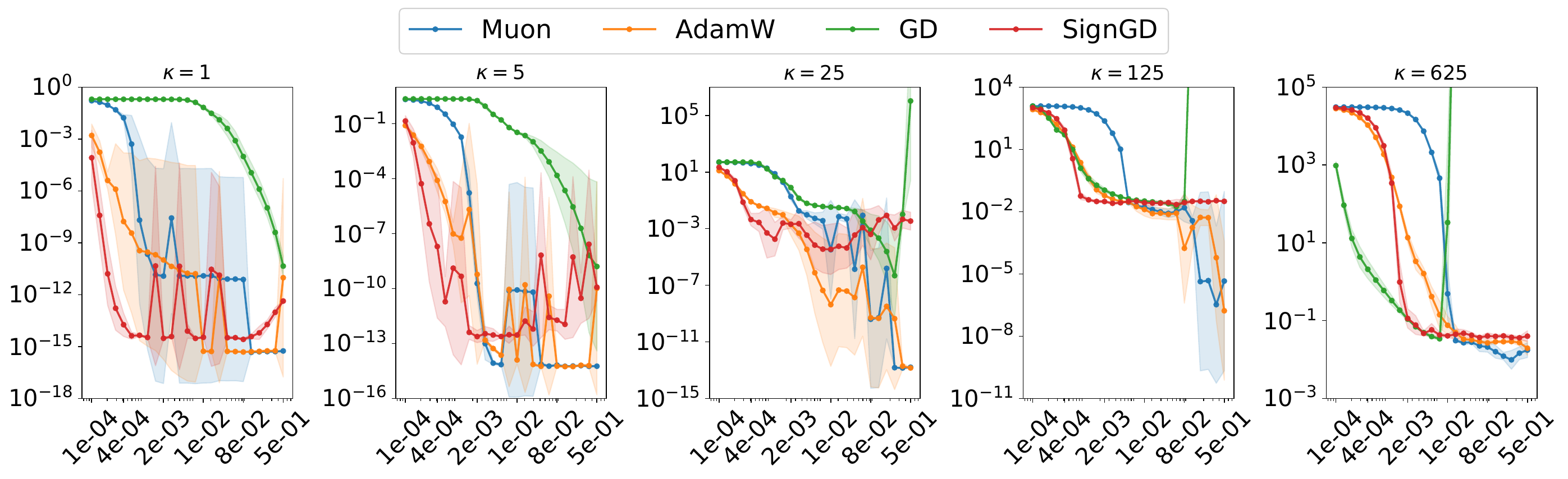}
\caption{\textbf{Matrix completion, conditioning sweep.} Each panel plots the
final reconstruction loss ($y$-axis, log scale; geometric mean over $3$ seeds
with a shaded $\pm$one-log-standard-deviation band) against the learning rate
($x$-axis, log scale), for condition number $\kappa$. Muon and AdamW are
indistinguishable in the well-conditioned regime; all methods degrade and
converge toward one another as $\kappa$ grows.}
\label{fig:compl-kappa}
\end{figure*}

\begin{figure*}[t]
\centering
\includegraphics[width=\textwidth]{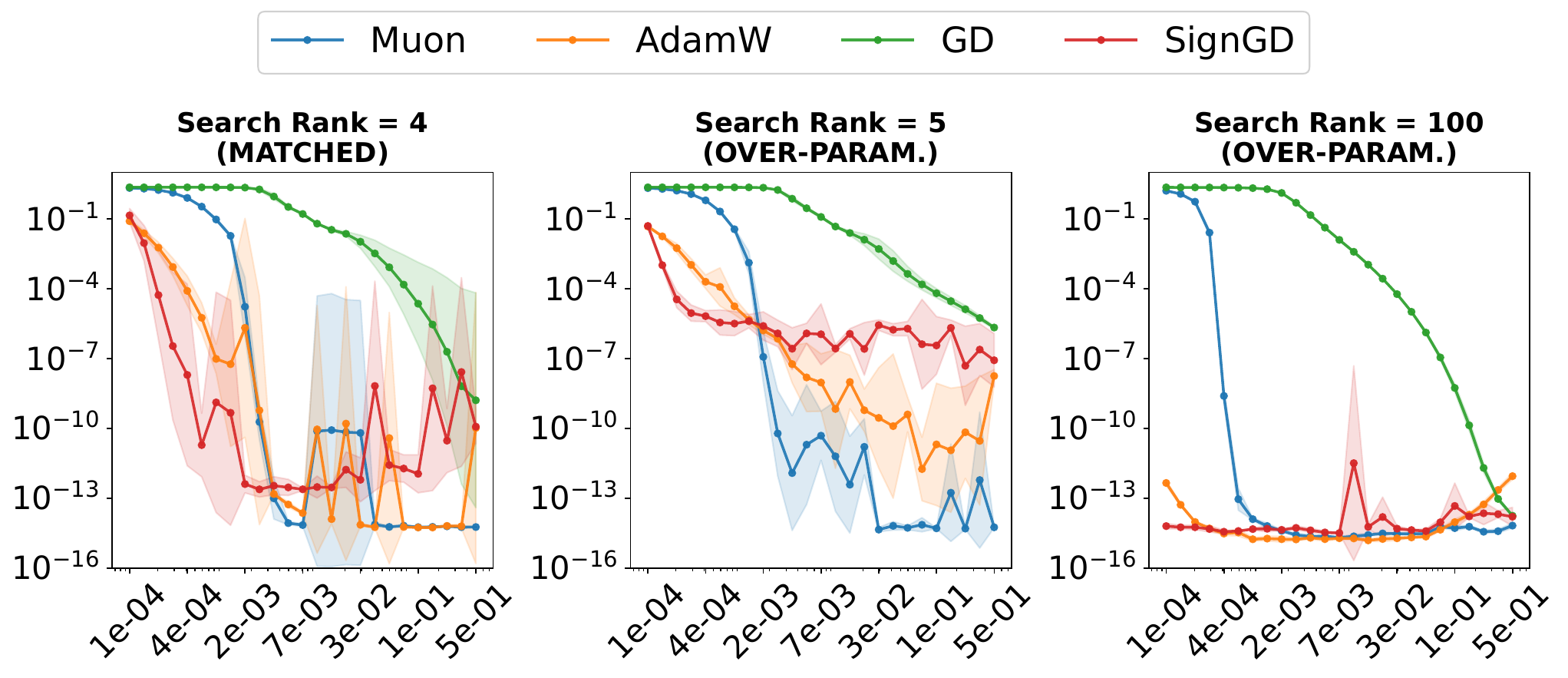}
\caption{\textbf{Matrix completion, search-rank sweep} (true rank $4$,
$\kappa{=}5$). Over-parameterization (rank $100$) widens the band of effective
learning rates and lets every method except GD reach machine precision.}
\label{fig:compl-rank}
\end{figure*}

\begin{figure*}[t]
  \centering
  \includegraphics[width=\textwidth]{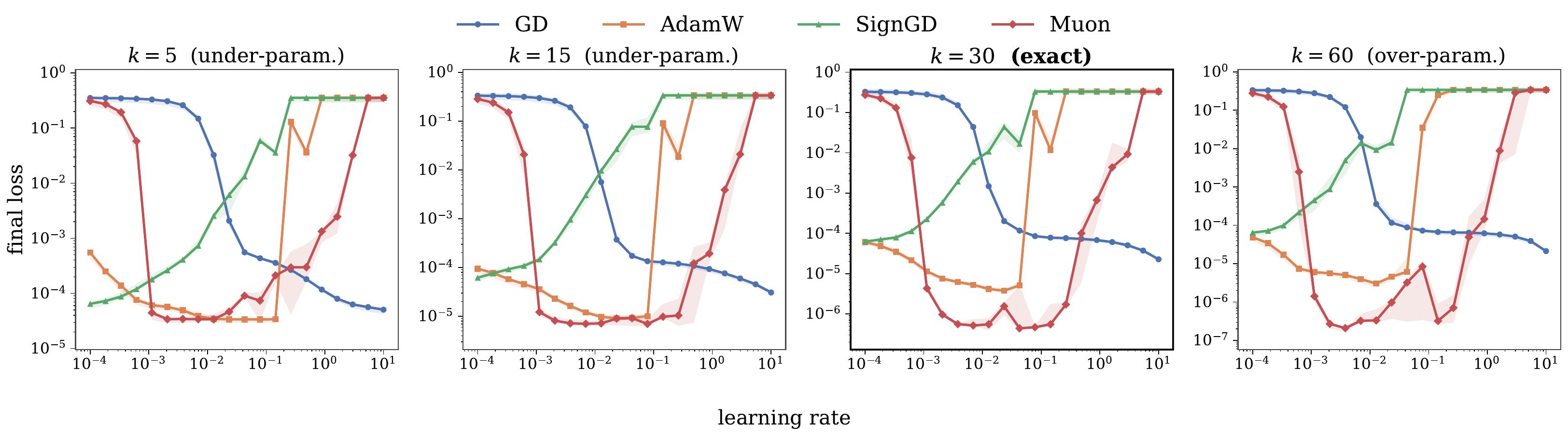}
  \caption{\textbf{Tensor-Train Noisy regime} (true rank $r^\star=30$, additive observation
    noise). Final loss versus learning rate for each optimizer, across search
    ranks $r\in\{5,15,30,60\}$ spanning under-, exactly- ($r=r^\star$, bold
    panel), and over-parameterized settings. Solid lines are the median over $3$
    seeds; shaded bands the min--max; the loss floors at the noise level rather
    than at zero.}
  \label{fig:tt-noisy}
\end{figure*}

\noindent \textbf{Implicit bias and generalization.}
A second line of work studies the implicit bias induced by \muon.
\citet{fan2025implicit} show that in linear classification, idealized \muon
converges to a solution that maximizes margin with respect to the spectral norm,
contrasting with the Euclidean and coordinate-wise biases of \texttt{SGD} and
\adam. Complementary empirical studies suggest that spectrum-aware updates can
improve generalization, particularly in imbalanced or long-tailed settings, by
promoting more uniform learning across principal components rather than focusing
on dominant directions \citep{vasudeva2025muon, wang2025muon, parviz2025nonlinear}. Related analyses
indicate that \muon yields more isotropic singular-value spectra than \adam and
may accelerate phenomena such as grokking in long-horizon training
\citep{zhang2025concurrence, wang2025muon, vasudeva2025muon, tveit2025muon, jha2026ream}.
Connections have also been drawn between \muon and second-order or preconditioned
methods, with several works interpreting its updates as approximations to
\shampoo \citep{gupta2018shampoo, jordan2024muon, shah2025practical}. Alternative
derivations and closely related formulations, some predating \muon, have appeared
from different theoretical viewpoints \citep{pethick2025training,
carlson2015preconditioned, lau2025polargrad, bernstein2024modular,
bernstein2024old, an2025asgo, sghaier2026probabilistic}.

\noindent \textbf{Preconditioning for matrix factorization.}
A separate body of work shows that preconditioning can dramatically accelerate
optimization in matrix factorization. In particular, \scaledgd and its variants
achieve linear convergence rates independent of the condition number under
appropriate initialization, in both exactly parameterized and over-parameterized
regimes \citep{tong2021accelerating, tong2021low, zhang2021preconditioned,
zhang2023preconditioned}, and these guarantees extend to small random
initialization \citep{xu2023power}. Closest to our motivation, recent work
analyzes the preconditioning effect of \muon's spectral orthogonalization
directly \citep{ma2026preconditioningbenefitsspectralorthogonalization}.

\noindent \textbf{Summary.}
Taken together, these results show that spectrum-aware and preconditioned methods
can offer strong theoretical and empirical benefits, but that their behavior
depends heavily on problem structure and initialization. This motivates a closer
examination of \muon in controlled settings such as matrix factorization, where
powerful alternatives already enjoy strong guarantees and where the advantage of
\muon over adaptive methods like \adamw is not immediately evident.

\section{Detailed experimental results}
\label{sec:app-results-detail}
 
This appendix expands the summary claims of Section~\ref{sec:exp-results} with the
specific numerical comparisons. All values refer to Table~\ref{tab:tuned}.
 
\paragraph{Horizontal basin shifts.}
Evaluating all methods at a common learning rate necessarily places at least one
of them far from its own optimum. This is visible as a horizontal shift between
each method's basin in every panel: for instance, GD's basin lies near the high
end of the grid while Muon's and AdamW's lie one to three decades lower.
 
\paragraph{Default-vs-tuned reordering.}
The ranking induced by a fixed default learning rate ($\approx 10^{-3}$) differs
from the tuned ranking in nearly all nineteen settings, and methods that look
worst at the default frequently become best after tuning. On low-rank
factorization GD is among the worst at the default rate yet attains the lowest
loss of all methods once tuned (Figure~\ref{fig:fact}); symmetrically, on both
NMF variants Muon is \emph{last} at the default rate but \emph{first} once tuned
(Figures~\ref{fig:nmf-combined}).
 
\paragraph{Effect of ill-conditioning.}
The effect of ill-conditioning is most dramatic for the non-adaptive methods on
factorization: GD, the best-tuned method in the well-conditioned regime
($1.6\times10^{-13}$ at $\kappa{=}1$), collapses to $1.3\times10^{-1}$ at
$\kappa{=}625$, where AdamW becomes best ($2.6\times10^{-7}$), Muon second
($9.4\times10^{-5}$), and SignGD diverges. Matrix completion shows the same
upward drift of the loss floor: at $\kappa{=}625$ all four methods are confined to
$O(10^{-2})$ and become nearly indistinguishable.
 
\paragraph{Problem-dependence of the tuned winner.}
Under per-optimizer tuning there is no universally dominant method. On
\emph{plain low-rank factorization}, AdamW and GD drive the loss to
$10^{-13}$--$10^{-9}$ across the whole conditioning range, while the Muon
configuration plateaus several orders higher ($10^{-8}$--$10^{-5}$). On
\emph{matrix completion}, Muon and AdamW are both reaching $10^{-16}$--$10^{-13}$ in the well-conditioned regime, and
over-parameterization (search rank $100$) lets all methods except GD reach machine
precision (Figure~\ref{fig:compl-rank}). On \emph{NMF}, the picture reverses: Muon
is the clear winner under both spectra, reaching $10^{-9}$--$10^{-6}$ while AdamW
stalls near $10^{-3}$ (with the single exception of the uniform-spectrum rank-$50$
case, where AdamW matches Muon at $\sim 4\times10^{-9}$) and GD/SignGD fail to
fit. Whichever optimizer a single-configuration study would crown thus depends
entirely on the problem and the learning rate chosen.

\section{Dynamics of spectral subspace recovery}
\label{sec:exp-dynamics}
The aggregate losses above tell us \emph{whether} a method fits the target but not
\emph{how} it does so. To probe the mechanism, we factorize a fixed
Gaussian-kernel target $K \in \mathbb{R}^{100 \times 100}$, symmetric positive
semidefinite, with eigendecomposition $K = Q\Lambda Q^{\top}$ and eigenvectors
ordered by descending eigenvalue, as $\hat{K} = UV^{\top}$ at full search rank
$k = 100$, training AdamW and Muon each at its own tuned learning rate. Every
$500$ iterations we take the left singular vectors $\{\hat{u}_i\}$ of the current
reconstruction $UV^{\top}$ and form the \emph{alignment matrix}
\begin{equation}
  M_{ij} \;=\; \bigl|\langle \hat{u}_i,\; q_j \rangle\bigr|,
  \label{eq:alignment}
\end{equation}
the absolute overlap between the $i$-th learned singular direction and the $j$-th
true eigenvector. A faithful, correctly ordered recovery of the eigenbasis yields
$M \to I$ (a sharp diagonal), whereas off-diagonal or diffuse mass indicates that
the learned subspace mixes or reorders eigendirections.
Figure~\ref{fig:dynamics} tracks $M$ over training. AdamW sharpens toward a clean
diagonal quickly and monotonically, recovering the eigenvectors faithfully and in
order, while Muon's alignment remains comparatively diffuse, with persistent
off-diagonal mass and slower diagonalization. This provides a mechanistic
explanation for the loss-level gap of
Sections~\ref{sec:exp-results}--\ref{sec:exp-spectrum}: on these structured and
ill-conditioned spectra, AdamW's per-coordinate adaptation locks onto the
dominant eigendirections in the right order, whereas Muon's orthogonalized
updates distribute capacity more evenly across directions and recover the
eigenbasis less cleanly. The heatmaps shown are for a representative tuned run;
the qualitative pattern is consistent across seeds.

\section{Spectrum families.}
\label{sec:app-spectral-profiles}
Across all experiments we fix $s_{\min}=10^{-3}$ and $s_{\max}=10$, so that the
condition number is $\kappa=10^4$. The target is $M=U\,\mathrm{diag}(s)\,V^\top
\in\mathbb{R}^{m\times n}$ with $m=n=100$ and $U,V$ drawn from the $Q$-factor of
i.i.d.\ Gaussian matrices; the spectrum $s\in\mathbb{R}^r_{>0}$ therefore has
length equal to the factorization rank $r=5$, and its entries are singular
values rather than eigenvalues. 

\emph{Common post-processing.} Every family below is generated, then sorted in
descending order, and then has its endpoints overwritten exactly:
$s_1:=s_{\max}$ and $s_r:=s_{\min}$. Consequently all families realize
$\kappa=10^4$ by construction, even where the sampled bulk does not reach the
endpoints; for \texttt{gaussian}, \texttt{u\_shaped}, and \texttt{geometric\_0.9}
the two extreme values should be read as planted outliers relative to the bulk.
Stochastic families are resampled per seed; \texttt{flat\_max},
\texttt{flat\_min}, \texttt{geometric}, \texttt{geometric\_<q>}, and
\texttt{powerlaw} are deterministic given $r$.

\begin{itemize}[leftmargin=*, itemsep=0.25em]
\item \texttt{flat\_max}: $s_r=s_{\min}$ and $s_1=\cdots=s_{r-1}=s_{\max}$
  (a single small singular value; mass at the top).
\item \texttt{flat\_min}: $s_1=s_{\max}$ and $s_2=\cdots=s_r=s_{\min}$
  (a single large singular value; mass at the bottom).
\item \texttt{uniform}: i.i.d.\ $\mathrm{Unif}[s_{\min},s_{\max}]$.
\item \texttt{log\_uniform}: $\tilde s_i = s_{\min}(s_{\max}/s_{\min})^{u_i}$
  with $u_i\sim\mathrm{Unif}[0,1]$, i.e.\ log-uniform across the four decades.
\item \texttt{geometric}: the deterministic geometric progression spanning both
  endpoints, $\tilde s_i = s_{\max}\,\rho^{\,i-1}$ with
  $\rho=(s_{\min}/s_{\max})^{1/(r-1)}$ ($\rho\approx0.829$ at $r=50$).
\item \texttt{geometric\_<q>}, $q\in\{0.3,0.5,0.9\}$: a deterministic geometric
  progression with \emph{fixed} ratio, clipped below at $s_{\min}$:
  $\tilde s_i=\max\!\big(s_{\min},\,s_{\max}q^{\,i-1}\big)$. Note that the
  resulting shape depends strongly on $q$ relative to $r$: at $r=50$ the clip
  binds for $84\%$ of the entries when $q=0.3$ and $72\%$ when $q=0.5$
  (yielding a large plateau at $s_{\min}$), whereas for $q=0.9$ it never binds
  ($s_{\max}q^{49}\approx5.7\cdot10^{-2}$) and the smallest value is supplied
  solely by the endpoint enforcement.
\item \texttt{powerlaw} ($\alpha=1$): the sequence $i^{-\alpha}$, $i=1,\dots,r$,
  is min--max normalized to $[0,1]$ and \emph{reflected} before rescaling,
  $\tilde s_i=s_{\min}+(s_{\max}-s_{\min})\big(1-\widetilde{i^{-\alpha}}\big)$.
  Despite the name this is not a heavy-tailed spectrum: at $\alpha=1$, $r=50$,
  $84\%$ of the values lie above $0.9\,s_{\max}$, making it close to
  \texttt{flat\_max} in shape.
\item \texttt{linear\_decay\_to\_smax}: $\tilde s = s_{\max}-(s_{\max}-s_{\min})\sqrt{u}$
  with $u\sim\mathrm{Unif}[0,1]$, a density that decreases linearly toward
  $s_{\max}$ and hence concentrates near $s_{\min}$.
\item \texttt{linear\_decay\_faster}: $\tilde s = s_{\min}+2-2\sqrt{u}$ with
  $u\sim\mathrm{Unif}[0,1]$. The additive constant is absolute rather than
  proportional to the range, so at $s_{\max}=10$ the sampled support is
  $[s_{\min},\,s_{\min}+2]$ and the largest value comes from the endpoint
  enforcement alone.
\item \texttt{gaussian\_<k>}, $k\in\{2,3\}$ (\texttt{gaussian} $\equiv k=3$):
  a truncated normal on $[s_{\min},s_{\max}]$. With
  $\mathrm{mid}:=(s_{\min}+s_{\max})/2$ and $\sigma:=(s_{\max}-s_{\min})/(2k)$,
  we draw $z_i\sim\mathcal N(0,1)$ conditioned on $|z_i|\le k$ (by rejection
  sampling, not clipping, so no atoms are placed at $\pm k$) and set
  $\tilde s_i=\mathrm{mid}+\sigma z_i$. The choice of $\sigma$ makes the
  truncation limits coincide with $[s_{\min},s_{\max}]$.
\item \texttt{u\_shaped\_<a>}: symmetric $\mathrm{Beta}(a,a)$ on $[0,1]$ mapped
  affinely to $[s_{\min},s_{\max}]$, $\tilde s_i=s_{\min}+(s_{\max}-s_{\min})z_i$
  with $z_i\sim\mathrm{Beta}(a,a)$ and $0<a<1$. We use the presets
  $a=0.9$ (\texttt{weak}), $a=0.5$ (\texttt{medium}, the default
  \texttt{u\_shaped}), and $a=0.2$ (\texttt{strong}); smaller $a$ concentrates
  more mass near both endpoints.
\item \texttt{spikes\_2\_<pmin>\_<pmax>}, with
  $(p_{\min},p_{\max})\in\{(0.99,0.01),(0.95,0.05),(0.9,0.1)\}$: a two-point
  spectrum placing $\lceil p_{\min}r\rfloor$ values at $s_{\min}$ and
  $\lceil p_{\max}r\rfloor$ at $s_{\max}$, interpolating between
  \texttt{flat\_min} and \texttt{flat\_max}. Since $p_{\min}+p_{\max}=1$ in all
  three cases, no residual mass remains to be allocated.
\end{itemize}

\section{Tensor-train regimes and parameterization sweep}
\label{sec:app-tt-detail}
 
This appendix gives the full specification of the regimes and parameterization
sweep summarized in Section~\ref{sec:problem-tt}.
 
\paragraph{Regimes.}
We study \eqref{eq:tt-objective} in two regimes. In the \emph{noiseless} regime
the target has true rank $r^\star = 4$ and is recovered exactly, so the attainable
loss is limited only by the optimizer's convergence; in the \emph{noisy} regime
the target has true rank $r^\star = 30$ and is corrupted by additive observation
noise, so the loss is floored at the statistical noise level rather than driven to
zero. The two regimes therefore probe complementary phenomena: optimization
geometry in the well-specified case and robustness to a nonzero residual in the
misspecified case.
 
\paragraph{Parameterization sweep.}
Within each regime we sweep the search rank, equivalently the TT-rank $r$ of the
factor $\mU$, across under-, exactly-, and over-parameterized values:
$r \in \{2, 4, 8, 20\}$ for the noiseless target (true rank $r^\star = 4$) and
$r \in \{5, 15, 30, 60\}$ for the noisy target (true rank $r^\star = 30$), with
$r = r^\star$ the exactly parameterized boundary case. Under-parameterization
($r < r^\star$) caps the attainable loss because $\mU\mU^\top$ cannot represent
$\mM^\star$, while over-parameterization ($r > r^\star$) enlarges the factor space
and introduces additional flat directions in the landscape.

\section{Relation to deep learning models}
\label{sec:tt-deep-learning}
The factorized objective~\eqref{eq:tt-objective} is not only a tensor-train
problem in its own right; it is the minimal instance of the product
parameterizations that pervade deep learning, which is what makes the optimizer
behavior we study here relevant beyond low-rank recovery.

\paragraph{Factorization as a shallow linear network.}
The map $\mU \mapsto \mU\mU^\top$ (or $(\mU,\mV)\mapsto\mU\mV^\top$ in the
asymmetric case of Appendix~\ref{sec:problem-mf-variants}) is exactly a two-layer
\emph{linear} network with no intervening nonlinearity: the factors are the
layer weights, and the product is the end-to-end map. Consequently the loss is
nonconvex in the factors despite being convex in the product, and the search
rank, equivalently the TT-rank, plays the role of the network width. The
over-parameterized regime $r > r^\star$ is precisely the width-overparameterized
regime of modern networks, and the flat directions it introduces in the
landscape are the source of the implicit bias of gradient methods studied for
matrix and deep factorizations \citep{gunasekar2017implicit,
arora2019implicit}. This is why over-parameterized factorization is a standard
proxy for the optimization of deep models: it isolates the effect of redundant
parameters on the trajectory while remaining analytically tractable.

\paragraph{Depth as a longer tensor-train chain.}
Increasing the order of the tensor-train chain, contracting $L$ cores
$\cG_1,\dots,\cG_L$ rather than two, is the algebraic analogue of increasing the
depth of a linear network, whose end-to-end map factorizes as
$\mW = \mW_L \mW_{L-1}\cdots\mW_1$. The matrix case~\eqref{eq:tt-contraction}
($L=2$) is the shallowest such chain, and depth introduces exactly the
ill-conditioning and balancing phenomena that motivate our conditioning sweep:
the product of many factors amplifies spectral imbalance, and the
overparameterized invariance group acts on each internal bond. Depth has been
shown to act as an implicit preconditioner that accelerates gradient descent on
these product objectives \citep{arora2018optimization}, so the sensitivity of
each optimizer to conditioning at $L=2$ is the base case of a phenomenon that
sharpens with depth.

\paragraph{Tensor-train structure inside trained networks.}
Beyond serving as a proxy, tensor-train factorizations appear directly inside
deep models. Reshaping a dense weight matrix into a high-order tensor and
representing it in TT format compresses fully connected and embedding layers by
orders of magnitude while preserving accuracy \citep{novikov2015tensorizing}, and
specific architectures correspond to specific tensor decompositions: recurrent
networks realize the tensor-train / matrix-product structure
\citep{khrulkov2018expressive}, while convolutional arithmetic circuits realize
the hierarchical Tucker decomposition \citep{cohen2016expressive}. In all of
these the rank of the decomposition controls expressivity exactly as the TT-rank
$r$ controls the capacity of~\eqref{eq:tt-objective}, so the interaction between
search rank, conditioning, and optimizer that we characterize on the factorized
problem speaks directly to how these layers are trained.

\section{Optimizer Comparison Across Tensor-Train Depth}
\label{sec:tt-depth-summary}

We benchmark four optimizers—Muon, AdamW, gradient descent (GD), and SignGD—on non-negative tensor-train (TT) factorization. To systematically vary problem conditioning, we use the tensor-train depth (equivalently, the tensor order or number of TT cores) as the primary experimental factor. Each TT core is represented via a matrix unfolding, a necessary implementation detail because Muon’s Newton–Schulz orthogonalization operates only on matrix-valued gradients; retaining a core in its native third-order form would effectively reduce Muon to momentum SGD on that factor. We evaluate two target regimes: a clean, low-rank, noise-free tensor and a noisy, higher-rank tensor. The study spans depths (2)–(6), an overparameterized factorization rank, a five-point learning-rate sweep, and three random seeds, yielding 600 total runs. Performance is reported as the best-tuned final loss for each optimizer, defined as the lowest mean squared error (MSE) achieved across the learning-rate grid and averaged over seeds.

\begin{figure}[t]
  \centering
  \includegraphics[width=\linewidth]{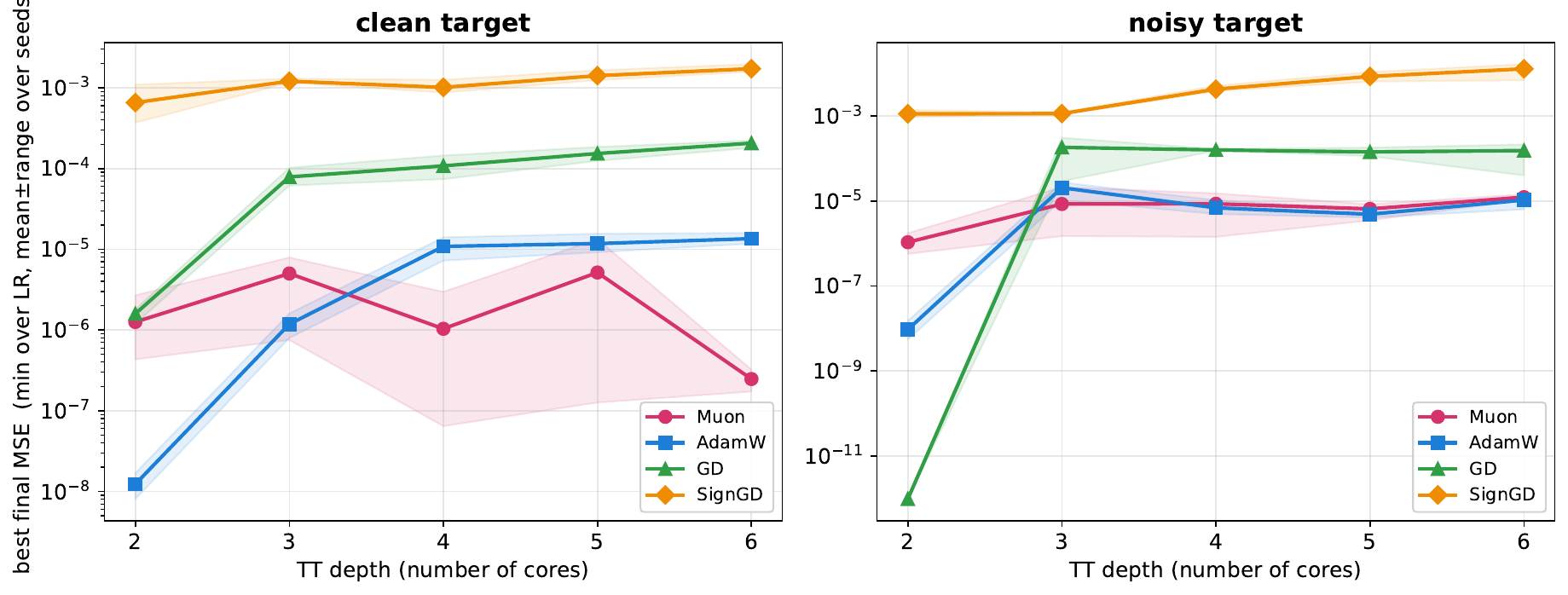}
  \caption{\textbf{Best-tuned reconstruction loss versus tensor-train depth.}
  Each curve shows the final MSE of one optimizer at its best learning rate
  (minimum over the grid) for a given depth, averaged over three seeds; shaded
  bands span the per-seed min--max. \textbf{Left:} clean, low-rank target.
  \textbf{Right:} noisy, higher-rank target. Search bond rank is fixed to the
  over-parameterized regime ($R{=}12$ clean, $R{=}30$ noisy). Lower is better;
  the $y$-axis is logarithmic.}
  \label{fig:tt-depth-loss}
\end{figure}

The results differ markedly between the two regimes. For the clean target, optimizer rankings change substantially with depth. AdamW achieves the lowest losses at shallow depths, while Muon performs comparatively poorly. However, beginning at depth (4), the performance of AdamW, GD, and SignGD deteriorates as the factorization deepens, whereas Muon maintains consistently low error. By depth (6), Muon outperforms the competing methods by approximately one to two orders of magnitude (Figure~\ref{fig:tt-depth-loss}, left). In contrast, performance on the noisy target remains relatively flat across depths, with all optimizers achieving similar losses (Figure~\ref{fig:tt-depth-loss}, right). The added noise introduces an irreducible error floor that limits the optimization gains obtainable from improved conditioning, thereby diminishing Muon’s advantage.

These findings suggest that Muon’s benefits arise primarily from its ability to mitigate optimization difficulties associated with ill-conditioned deep factorizations, rather than from a general robustness advantage. At the same time, two limitations should be noted. First, increasing tensor-train depth changes the underlying optimization problem itself, so the observed trends reflect changes in relative optimizer performance across a family of increasingly difficult tasks rather than within a fixed loss landscape. Second, the reported results correspond only to the overparameterized setting considered here and may not fully characterize behavior at other scales or rank regimes.

\section{Softplus, Its Taylor Surrogate, and Nonlinear Matrix Factorization}
\label{app:softplus-mf}

Let $X\in\mathbb{R}^{N\times d}$ collect the $N$ inputs. We fit the Gaussian
kernel $K\in\mathbb{R}^{N\times N}$, $K_{ij}=\exp(-\lVert x_i-x_j\rVert^2/2\sigma^2)$,
with the model $\hat K=\phi(XW)\,V^\top$, where $W\in\mathbb{R}^{d\times R}$,
$V\in\mathbb{R}^{N\times R}$, the search rank $R$ is the hidden width, and $\phi$
acts entrywise. This is a \emph{nonlinear} matrix factorization: for $\phi=\mathrm{id}$
it collapses to the bilinear low-rank model $\hat K=X(WV^\top)$.

\paragraph{Softplus and its expansion.}
The softplus $\phi(z)=\log(1+e^{z})$ has derivative $\phi'=\sigma$ (the logistic
sigmoid). Since $\sigma-\tfrac12$ is odd, the Maclaurin series of $\phi$ keeps only
the constant, linear, and even-order terms,
\begin{equation}
\phi(z)=\log 2+\tfrac12 z+\tfrac18 z^{2}-\tfrac{1}{192}z^{4}+\mathcal{O}(z^{6}).
\end{equation}
The surrogate used in the experiments, $\tilde\phi(z)=\log2+\tfrac12 z+\tfrac18 z^{2}$,
is the truncation after the quadratic term (the cubic coefficient vanishes), and is
accurate when the pre-activations $XW$ are small, as enforced by the input scale.

\paragraph{The quadratic surrogate is an explicit feature factorization.}
With $A=\tilde\phi(XW)$, expanding term by term gives
\begin{equation}
\hat K = A V^\top
= \underbrace{\log2\,\mathbf{1}_{N\times R}V^\top}_{\text{constant }(\mathrm{rank}\le1)}
\;+\;\underbrace{\tfrac12\,(XW)V^\top}_{\mathrm{rank}\le\min(R,d)}
\;+\;\underbrace{\tfrac18\,(XW)^{\odot 2}V^\top}_{\text{degree-2 lift}},
\end{equation}
where $\odot$ is the Hadamard product. The linear term is exactly the classical
rank-$R$ bilinear factorization, whose rank is further capped by
$\operatorname{rank}(X)\le d$. The Hadamard-square term injects degree-2 monomial
features: each column of $XW$ lies in the $\le d$-dimensional column space
$\operatorname{col}(X)$, so the columns of $(XW)^{\odot 2}$ lie in its symmetric
square $\operatorname{Sym}^2(\operatorname{col}(X))$, of dimension
$\le\binom{d+1}{2}$. Hence the model factorizes $K$ through the fixed feature set
$\{\mathbf 1\}\cup\operatorname{col}(X)\cup\operatorname{Sym}^2(\operatorname{col}(X))$,
and
\begin{equation}
\operatorname{rank}(\hat K)\;\le\;\min\!\Big(R,\ \tbinom{d+2}{2}\Big).
\label{eq:rank-cap}
\end{equation}
The quadratic model therefore \emph{cannot exceed} rank $\binom{d+2}{2}$
($=21$ for $d=5$) no matter how large the search rank $R$ is.

\begin{figure*}[t]
  \centering
  \includegraphics[width=\linewidth]{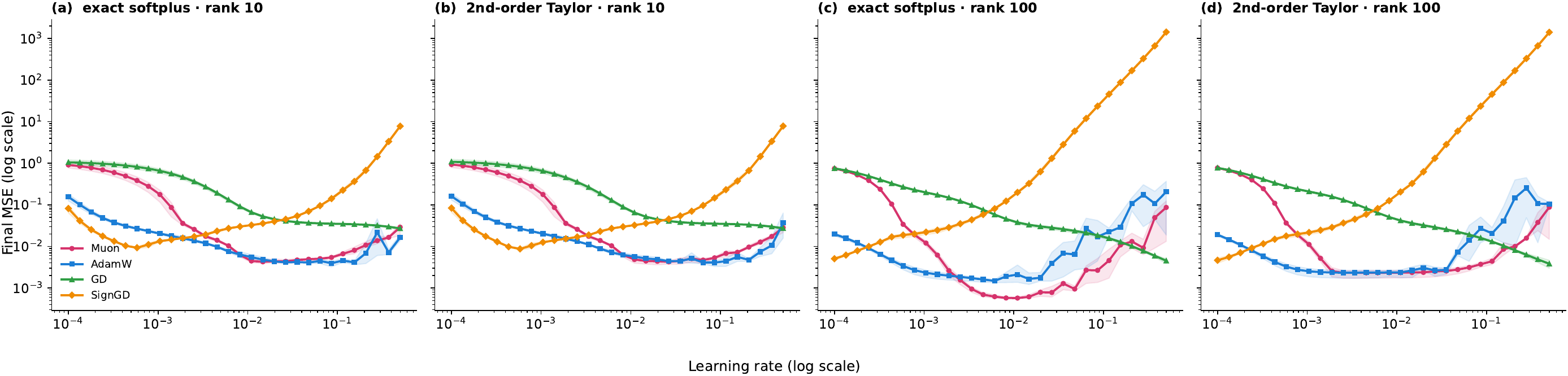}
  \caption{\textbf{Learning-rate stability across activation and capacity.} Final
  MSE (log) vs.\ learning rate (log) for four optimizers fitting the Gaussian
  kernel as $\hat K=\phi(XW)V^{\top}$; mean over three seeds, bands show min--max,
  divergent runs capped at $10^{5}$. The activation $\phi$ is exact softplus or its
  second-order Taylor surrogate, and $R$ is the search rank: \textbf{(a)} softplus,
  $R{=}10$; \textbf{(b)} Taylor, $R{=}10$; \textbf{(c)} softplus, $R{=}100$;
  \textbf{(d)} Taylor, $R{=}100$. At $R{=}10$ the rank bottleneck binds and the two
  activations match (a)$\approx$(b); at $R{=}100$ the degree-2 surrogate is rank-capped
  (Eq.~\ref{eq:rank-cap}) while exact softplus is not, so (c) reaches a lower error
  than (d). Muon attains the lowest loss over the widest stable range; SignGD
  diverges at large learning rates.}
  \label{fig:kernel-lr-sweep}
\end{figure*}

\paragraph{Consequence for the comparison.}
Equation~\eqref{eq:rank-cap} predicts a sharp regime split, which
Fig.~\ref{fig:kernel-lr-sweep} confirms. When $R\le\binom{d+2}{2}$ (here $R=10$),
the search rank is the binding constraint and the truncation is invisible: exact
softplus and its surrogate behave almost identically (panels~(a) vs~(b)). When
$R>\binom{d+2}{2}$ (here $R=100$), the degree-2 cap binds for the surrogate but not
for exact softplus, so the exact activation attains a strictly lower error
(panel~(c) below~(d)). In both regimes the activation is a mild perturbation of the
same nonlinear-factorization family, while the search rank $R$ governs how much of
the kernel's polynomial spectrum the model can represent, and, empirically, is what
separates the optimizers.

\section{Analytical Solutions of the Matrix Factorization for AdamW and Muon}
\label{app:proxy-derivations}

\subsection{Problem Setup}
\label{app:proxy-setup}

We consider the Matrix Factorization objective function:
\begin{equation}
    \mathcal{L}(U, V) = \frac{1}{2} \| U V^T - R \|_F^2,
\end{equation}
where $U, V \in \mathbb{R}^{N \times r}$. At iteration $t$, we seek the optimal update
step $\Delta U$ for the factor $U$ (the derivation for $V$ is symmetric).
Let $G_t = \nabla_U \mathcal{L} = (U_t V_t^T - R)V_t$ be the gradient at step $t$.

Since the global optimization is intractable, iterative optimizers solve a
\textbf{local proxy problem} at each step. We derive the analytical solution
for these proxy problems. Throughout, $v_t$ (lowercase) denotes the per-coordinate
second-moment estimate, with entries $v_{t,ij}$; this is distinct from the factor
$V_t$.

\subsection{The AdamW Analytical Solution}
\label{app:proxy-adamw}

\begin{proposition}
Under the decoupled--weight-decay approximation of Loshchilov \& Hutter (2017), the
AdamW update is the analytical solution to minimizing the linearized loss subject to
an adaptive Mahalanobis-distance trust region, with weight decay applied as a separate
additive step.
\end{proposition}

\paragraph{The Optimization Problem.}
We first consider the linearized loss penalized by the local curvature estimated by
the diagonal second-moment estimate $v_t$:
\begin{equation}
    \Delta U^{\dagger} = \argmin_{\Delta U \in \mathbb{R}^{N \times r}}
    \left(
        \underbrace{\langle G_t, \Delta U \rangle}_{\text{Linear Descent}}
        + \underbrace{\frac{1}{2\eta} \sum_{i,j} \sqrt{v_{t,ij}}\, (\Delta U_{ij})^2}_{\text{Adaptive Trust Region}}
    \right).
\end{equation}

\paragraph{Derivation of the trust-region step.}
The objective is separable across coordinates. Taking the derivative with respect to
a single entry $\Delta U_{ij}$ and setting it to zero,
\begin{equation}
    G_{t,ij} + \frac{1}{\eta} \sqrt{v_{t,ij}}\, \Delta U_{ij} = 0
    \quad\Longrightarrow\quad
    \Delta U^{\dagger}_{ij} = -\,\eta\, \frac{G_{t,ij}}{\sqrt{v_{t,ij}}}.
\end{equation}

\paragraph{Decoupled weight decay.}
AdamW then applies weight decay as a \emph{separate} step that is not passed through
the preconditioner:
\begin{equation}
    \boxed{\;\Delta U^*_{ij} = -\,\eta \left( \frac{G_{t,ij}}{\sqrt{v_{t,ij}}} + \lambda\, U_{t,ij} \right).\;}
\end{equation}

\paragraph{Remark (why decoupling matters).}
Had we instead folded the penalty $\tfrac{\lambda}{2}\|U_t + \Delta U\|_F^2$ directly
into the proxy, the exact stationary point would be
\begin{equation}
    \left( \frac{1}{\eta}\sqrt{v_{t,ij}} + \lambda \right) \Delta U_{ij}
    = -\big(G_{t,ij} + \lambda U_{t,ij}\big)
    \;\Longrightarrow\;
    \Delta U_{ij} = -\,\eta\,\frac{G_{t,ij} + \lambda U_{t,ij}}{\sqrt{v_{t,ij}} + \eta\lambda}.
\end{equation}
This \emph{couples} the decay to the preconditioner (dividing the decay term by
$\sqrt{v_{t,ij}}$) and is precisely Adam with $L_2$ regularization, \emph{not} AdamW.
AdamW is recovered by (i) neglecting $\eta\lambda$ in the denominator and (ii)
decoupling the decay so that $\lambda U_{t,ij}$ is \emph{not} rescaled by the
curvature. Hence the boxed expression is an approximation, not the exact minimizer of
the coupled proxy.

\paragraph{Conclusion.}
AdamW rescales every coordinate independently based on the diagonal curvature
$\sqrt{v_{t,ij}}$, and adds an unpreconditioned weight-decay term.


\subsection{The Muon Analytical Solution}
\label{app:proxy-muon}

\begin{proposition}
The Muon update is the exact analytical solution to the Orthogonal Procrustes Problem.
It finds the orthonormal-column update direction that aligns maximally with the
\emph{negative} gradient; this direction is the polar factor of $-G_t$.
\end{proposition}

\paragraph{The Optimization Problem.}
We seek the matrix $O \in \mathbb{R}^{N \times r}$ closest to the negative gradient
$-G_t$, constrained to have orthonormal columns:
\begin{equation}
    O^* = \argmin_{O\,:\,O^T O = I_r} \; \| (-G_t) - O \|_F^2 .
\end{equation}
The update is then $\Delta U = \eta\, O^*$.

\subsubsection{Step-by-Step Derivation}

\paragraph{Step 1: Expand the objective.}
\begin{align}
    \| -G_t - O \|_F^2
    &= \Tr\!\left( (G_t + O)^T (G_t + O) \right) \\
    &= \Tr(G_t^T G_t) + \Tr(O^T O) + 2\,\Tr(G_t^T O).
\end{align}
The term $\Tr(G_t^T G_t)$ is constant, and $\Tr(O^T O) = \Tr(I_r) = r$ is constant.
The cross term carries a \textbf{plus} sign, so minimizing the norm is equivalent to
\textbf{minimizing} $\Tr(G_t^T O)$ , equivalently maximizing the alignment with the
negative gradient, $\Tr\!\big((-G_t)^T O\big)$:
\begin{equation}
    \min_{O\,:\,O^T O = I_r} \; \Tr(G_t^T O).
\end{equation}

\paragraph{Step 2: Singular Value Decomposition.}
Let the (thin) SVD of the gradient be $G_t = P \Sigma Q^T$, where $P \in \mathbb{R}^{N \times r}$
has orthonormal columns ($P^T P = I_r$), $\Sigma \in \mathbb{R}^{r \times r}$ is diagonal
with singular values $\sigma_i \ge 0$, and $Q \in \mathbb{R}^{r \times r}$ is orthogonal.
Then
\begin{equation}
    \Tr(G_t^T O) = \Tr\!\big( Q \Sigma P^T O \big)
    = \Tr\!\big( \Sigma\, (P^T O Q) \big),
\end{equation}
using the cyclic property of the trace.

\paragraph{Step 3: Bound the matrix $Z$.}
Define $Z = P^T O Q \in \mathbb{R}^{r \times r}$. Each of $P$, $O$ has orthonormal
columns and $Q$ is orthogonal, so each is a (semi-)isometry with spectral norm $1$.
Hence $\sigma_{\max}(Z) \le \|P^T\|_2 \,\|O\|_2 \,\|Q\|_2 = 1$. (Note $Z$ need
\emph{not} be orthogonal when $N > r$, since $O O^T \ne I_N$; we only require the norm
bound.) In particular, every diagonal entry satisfies $|Z_{ii}| \le \sigma_{\max}(Z) \le 1$.
The objective becomes
\begin{equation}
    \Tr(\Sigma Z) = \sum_{i=1}^r \sigma_i Z_{ii}.
\end{equation}

\paragraph{Step 4: Minimize.}
Since $\sigma_i \ge 0$ and $|Z_{ii}| \le 1$, the sum $\sum_i \sigma_i Z_{ii}$ is
\textbf{minimized} when $Z_{ii} = -1$ for all $i$, i.e.\ $Z = -I_r$ (attained by the
admissible choice $O = -PQ^T$). Solving back,
\begin{align}
    P^T O Q &= -I_r \\
    O^* &= -\,P\,Q^T.
\end{align}
\begin{equation}
    \boxed{\;O^* = -\,P Q^T,
    \qquad
    \Delta U = \eta\, O^* = -\,\eta\, P Q^T.\;}
\end{equation}
Equivalently, $O^*$ is the polar factor of the \emph{negative} gradient $-G_t$, and the
update $\Delta U = -\eta\,PQ^T$ is a genuine descent step (it removes the singular-value
magnitudes of $G_t$, keeping only its $PQ^T$ ``direction'').

\subsubsection{Connection to Newton--Schulz}
The matrix $P Q^T$ is the \textbf{polar factor} of $G_t$. Computing the SVD at every
step is expensive, so Muon computes $PQ^T$ via the Newton--Schulz iteration
\begin{equation}
    X_{k+1} = \frac{1}{2} X_k \big(3I - X_k^T X_k\big),
    \qquad X_0 = \frac{G_t}{\|G_t\|_2},
\end{equation}
which converges quadratically to $P Q^T$. The descent update is then
$\Delta U = -\eta\, X_\infty = -\eta\, P Q^T$, consistent with the boxed solution above.

\subsection{Implications for Matrix Factorization}
\label{app:proxy-implications}

This derivation highlights the fundamental difference in update scaling. Both updates
act on matrices in $\mathbb{R}^{N \times r}$.

\paragraph{1. AdamW scaling (dimension dependent).}
Assuming active gradients ($G_{ij}/\sqrt{v_{ij}} \approx \pm 1$) and negligible weight
decay, the squared norm of the update scales with the number of entries $N \times r$:
\begin{equation}
    \| \Delta W_{\text{Adam}} \|_F^2 \approx \sum_{i,j} \eta^2 = \eta^2 (N \cdot r).
\end{equation}

\paragraph{2. Muon scaling (rank dependent).}
Since $O^* = -P Q^T$ has orthonormal columns, $(O^*)^T O^* = Q P^T P Q^T = I_r$, so the
sign is immaterial to the norm:
\begin{equation}
    \| \Delta W_{\text{Muon}} \|_F^2 = \eta^2 \,\Tr\!\big( (O^*)^T O^* \big)
    = \eta^2 \,\Tr(I_r) = \eta^2 r.
\end{equation}

\paragraph{Conclusion.}
\begin{equation}
    \frac{\| \Delta W_{\text{Adam}} \|_F}{\| \Delta W_{\text{Muon}} \|_F} \approx \sqrt{N}.
\end{equation}
In our experiments with $N = 100$, AdamW naturally takes steps $\approx 31.6$ times
larger than Muon for the same learning rate $\eta$. This explains the necessity of
scaling $\eta_{\text{Muon}} \approx \sqrt{N}\,\cdot\,\eta_{\text{AdamW}}$ to achieve
comparable convergence rates.

\end{document}